\documentclass[pdflatex,sn-mathphys-num]{sn-jnl}

\usepackage{graphicx}%
\usepackage{multirow}%
\usepackage{amsmath,amssymb,amsfonts}%
\usepackage{amsthm}%
\usepackage{mathrsfs}%
\usepackage[title]{appendix}%
\usepackage{xcolor}%
\usepackage{textcomp}%
\usepackage{manyfoot}%
\usepackage{booktabs}%
\usepackage{algorithm}%
\usepackage{algorithmicx}%
\usepackage{algpseudocode}%
\usepackage{listings}%
\usepackage[caption=false,font=footnotesize]{subfig}
\usepackage{ragged2e}
\usepackage{array}
\newcolumntype{P}[1]{>{\centering\arraybackslash}p{#1}}

\usepackage{tabulary}
\newcolumntype{K}[1]{>{\centering\arraybackslash}p{#1}}

\newcommand{\new}[1]{#1} 

\begin{document}

\title[TGIF2: Extended Text-Guided Inpainting Forgery Dataset \& Benchmark]{TGIF2: Extended Text-Guided Inpainting Forgery Dataset \& Benchmark}

\author*[1]{\fnm{Hannes} \sur{Mareen}}\email{hannes.mareen@ugent.be}
\author[2]{\fnm{Dimitrios} \sur{Karageorgiou}}\email{dkarageo@iti.gr}
\author[2]{\fnm{Paschalis} \sur{Giakoumoglou}}\email{giakoupg@iti.gr}
\author[1]{\fnm{Peter} \sur{Lambert}}\email{peter.lambert@ugent.be}
\author[2]{\fnm{Symeon} \sur{Papadopoulos}}\email{papadop@iti.gr}
\author[1]{\fnm{Glenn} \spfx{Van} \sur{Wallendael}}\email{glenn.vanwallendael@ugent.be}

\affil[1]{\orgdiv{IDLab}, \orgname{Ghent University -- imec}, \orgaddress{\city{Gent}, \country{Belgium}}}

\affil[2]{\orgdiv{Information Technologies Institute}, \orgname{CERTH}, \orgaddress{\city{Thessaloniki}, \country{Greece}}}

\abstract{
Generative AI has made text-guided inpainting a powerful image editing tool, but at the same time a growing challenge for media forensics. Existing benchmarks, including our text-guided inpainting forgery (TGIF) dataset, show that image forgery localization (IFL) methods can localize manipulations in spliced images but struggle not in fully regenerated (FR) images, while synthetic image detection (SID) methods can detect fully regenerated images but cannot perform localization.  
With new generative inpainting models emerging and the open problem of localization in FR images remaining, updated datasets and benchmarks are needed.  
\new{We introduce TGIF2, an extended version of TGIF, that captures recent advances in text-guided inpainting and enables a deeper analysis of forensic robustness. TGIF2 augments the original dataset with edits generated by FLUX.1 models, as well as with random non-semantic masks. Using the TGIF2 dataset, we conduct a forensic evaluation spanning IFL and SID, including fine-tuning IFL methods on FR images and generative super-resolution attacks.}
\new{Our experiments show that both IFL and SID methods degrade on FLUX.1 manipulations, highlighting limited generalization. Additionally, while fine-tuning improves localization on FR images, evaluation with random non-semantic masks reveals object bias. Furthermore, generative super-resolution significantly weakens forensic traces, demonstrating that common image enhancement operations can undermine current forensic pipelines.}
In summary, TGIF2 provides an updated dataset and benchmark, \new{which enables} new insights into the challenges posed by modern inpainting and AI-based image enhancements. TGIF2 is available at \url{https://github.com/IDLabMedia/tgif-dataset}.
}


\maketitle

\section{Introduction}\label{sec:intro}
Digital image manipulation has become increasingly accessible and efficient, producing more realistic forged images with recent advances in generative AI (GenAI) technology~\cite{rombach2022sd}. A few years ago, creating convincing image edits required expertise in software such as Adobe Photoshop and significant manual effort. More recently, deepfake technology~\cite{thies2016face2face} enabled realistic face swaps and modifications, but these were typically limited to faces and required technical skills. Today, modern open-source and commercial GenAI models, such as Stable Diffusion (SD)~\cite{rombach2022sd}, Adobe Firefly~\cite{adobe2023firefly}, and FLUX~\cite{flux2024}, can perform high-resolution edits that are often indistinguishable from authentic images. Additionally, these GenAI models are accessible to users without technical expertise as they work by simple text prompting~\cite{nichol2021glide, rombach2022sd}. Such capabilities have positive applications in creative industries, image restoration, and education. However, they also pose serious risks, including fraud, disinformation, and the creation of fake evidence~\cite{verdoliva2020media}.

\begin{figure*}[h]
\centering
\subfloat[Original]{\includegraphics[width=0.24\linewidth]{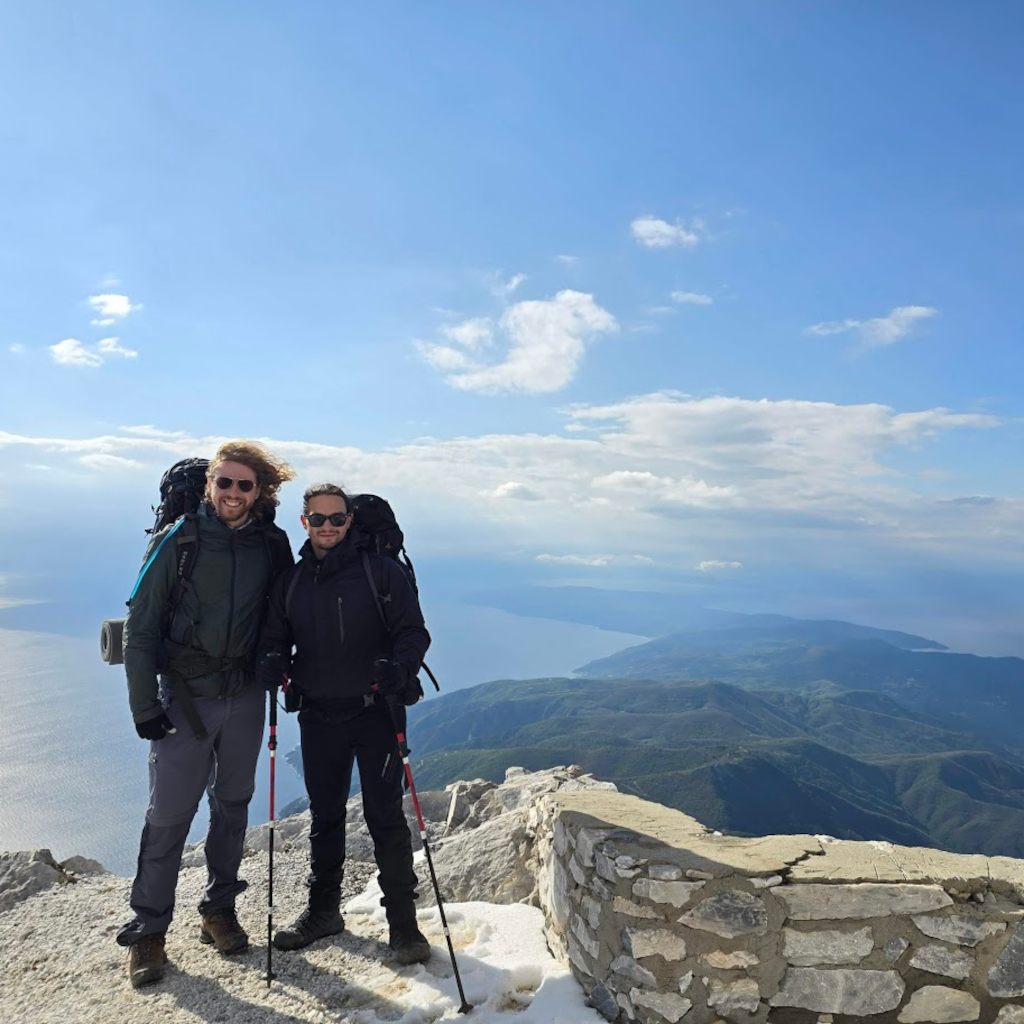}
\label{fig:intro-orig}}
\hfill
\subfloat[Mask]{\includegraphics[width=0.24\linewidth]{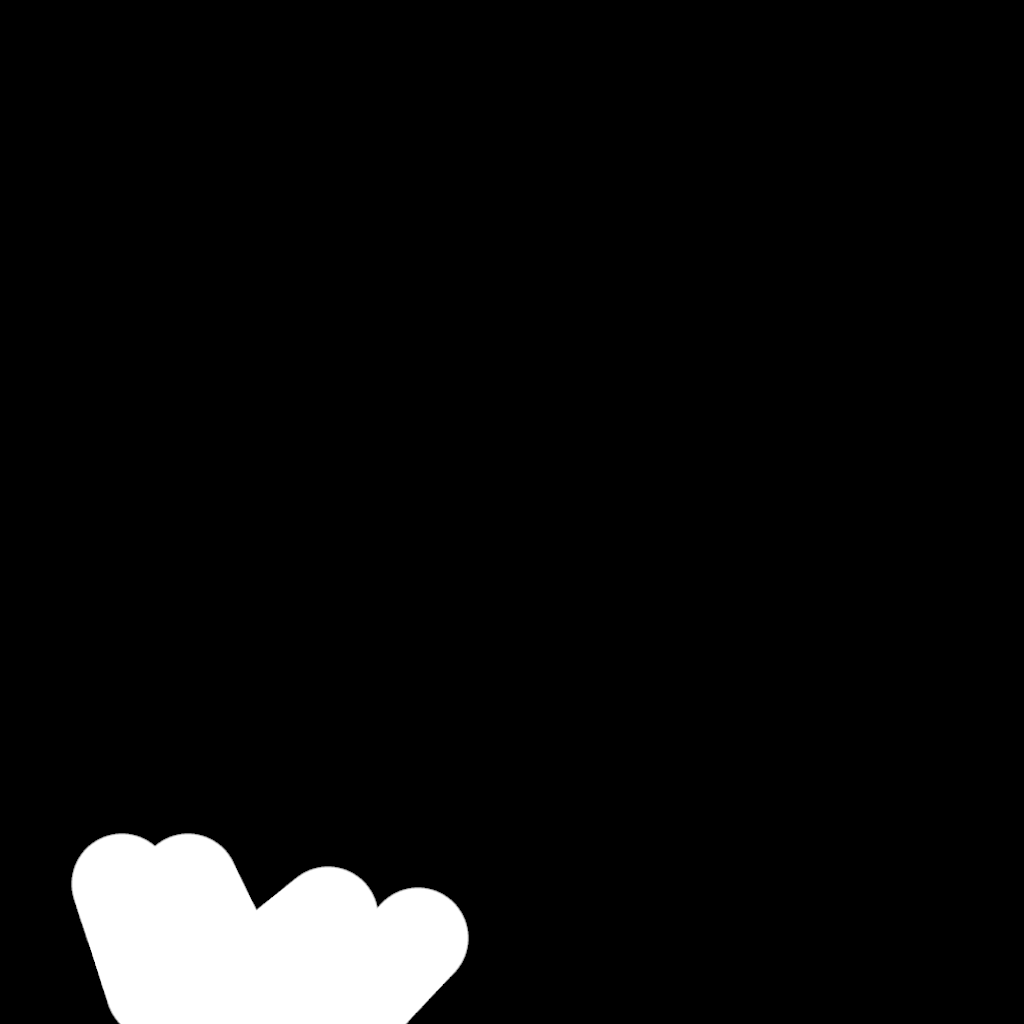}
\label{fig:intro-mask}}
\hfill
\subfloat[Fully regenerated]{\includegraphics[width=0.24\linewidth]{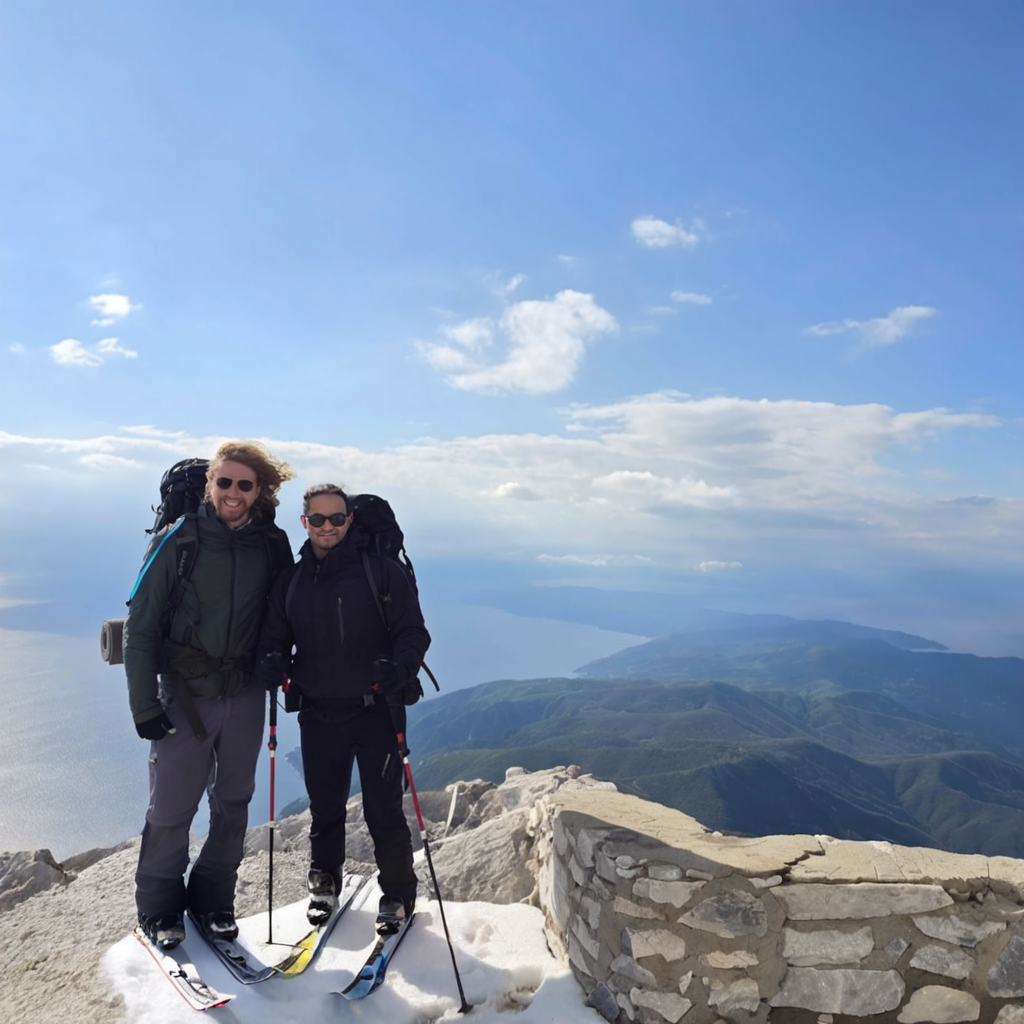}
\label{fig:intro-fr}}
\hfill
\subfloat[Spliced]{\includegraphics[width=0.24\linewidth]{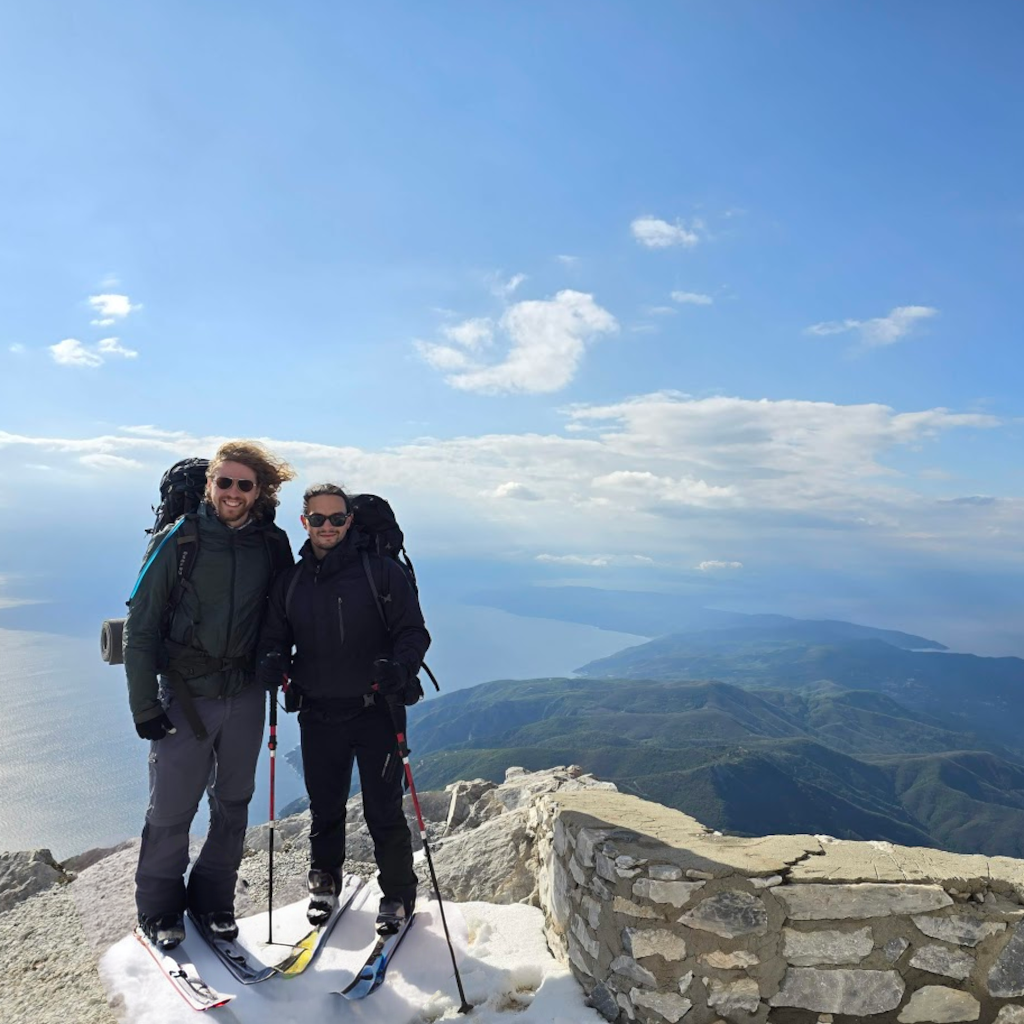}
\label{fig:intro-sp}}
\\
\subfloat[Fully regenerated (zoom)]{\includegraphics[width=0.49\linewidth]{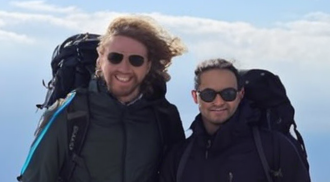}
\label{fig:intro-fr-zoom}}
\hfill
\subfloat[Original \& Spliced (zoom)]{\includegraphics[width=0.490\linewidth]{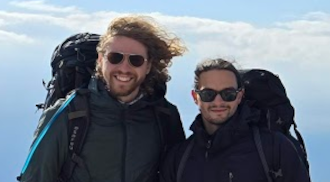}
\label{fig:intro-sp-zoom}}

\caption{Examples of the two ways of inpainting. An (a) authentic image can be inpainted in a (b) selected region as mask, with \emph{skis} as prompt. In the GenAI-based inpainting process (here using SDXL), (c) the full image is regenerated during editing. To minimize artifacts, (d) only the region corresponding to the mask can be spliced into the authentic image. (e) and (f) provide zoomed-in versions to more clearly show the differences between the original or spliced pixels, on the one hand, and the regenerated pixels outside of the masked area, on the other. Subtle differences due to the regenerative process can be observed in the teeth, sunglasses, beard, etc.
\label{fig:intro}}
\end{figure*}

Among GenAI approaches, text-guided inpainting~\cite{nichol2021glide, rombach2022sd} is particularly powerful. It allows users to select a region of an (authentic) image and provide a textual prompt describing the desired edit. Diffusion-based models then regenerate the image so that the edited region matches the prompt, while leaving the rest of the image visually intact. Depending on the workflow, the edited region may be spliced into the original image (i.e., only changing pixels in the selected region) or the model may regenerate the entire image (i.e., potentially changing all pixels in the image, yet only semantically altering the selected region).
This is demonstrated in Fig.~\ref{fig:intro}.
In previous work~\cite{mareen2024tgif}, we showed that when regenerating the entire image during editing, the forensic traces that traditional image forgery localization (IFL) methods use are destroyed. Moreover, our previous work has found that synthetic image detection (SID) methods can detect fully regenerated images as synthetic, but they are not designed to localize manipulated regions.

The Text-Guided Inpainting Forgery (TGIF) dataset proposed in our previous work~\cite{mareen2024tgif} provided high-resolution images manipulated with three text-guided inpainting models: SD2~\cite{rombach2022sd}, SDXL~\cite{podell2023sdxl}, and Adobe Firefly~\cite{adobe2023firefly}), including both spliced and fully regenerated images. While TGIF provided a valuable dataset, benchmark and insights, recent developments in generative models require an updated, more comprehensive dataset.
First, new generative inpainting models were released since.
Most notably, FLUX.1~\cite{flux2024} is a new generation of diffusion model that replaces the traditional U-net and diffusion pipeline of traditional diffusion models (such as SD) with a transformer and flow-matching framework. As such, FLUX.1 achieves higher-quality local and global generations~\cite{imgsys}.
Second, the previously acquired insight that generative edits significantly impact IFL models requires further exploration for potential solutions.
This highlights the need for a more comprehensive benchmark and experiments that deepen our understanding of the impact of recent GenAI inpainting models on forensic methods.

To this end, we introduce TGIF2, an extended version of TGIF, with several new contributions:
\begin{itemize}
\item \textbf{FLUX inpainting:} \new{TGIF was limited to using SD2, SDXL, and Adobe Firefly to inpaint images. TGIF2 extends the TGIF dataset by additionally including FLUX.1 models ([schnell], [dev] \& Fill [dev]}). The FLUX.1 models are a new generation of diffusion models, and are among the most capable open-source text-guided inpainting models available today~\cite{imgsys}. In contrast to traditional SD models, they utilize flow-matching instead of denoising diffusion and rely on large multimodal transformers rather than a U-net for better contextual reasoning.
\item \textbf{Random non-semantic masks:} In addition to semantic, object-centric masks, TGIF2 includes randomly positioned, non-semantic masks, allowing us to evaluate potential biases of IFL methods toward objects or other semantically meaningful regions.
\item \new{\textbf{New experimental contributions:} We provide new results for IFL methods fine-tuned on TGIF2's fully regenerated images, the relation of the IFL performance with generative quality, and the effect of super resolution (SR)~\cite{wang2021realesrgan} as an attack against IFL and SID methods} 
\end{itemize}

These additions make TGIF2 not only a larger dataset but enable new insights into forensic challenges.
\new{
First, the inclusion of FLUX.1 inpainted images allows evaluation of how current IFL and SID methods generalize to newer generative models.
Second, the provided random, non-semantic masks may reveal potential biases of IFL methods toward objects or semantically meaningful regions, highlighting limitations in their robustness. For example, we demonstrate that IFL models that are fine-tuned on non-random, semantic FR data enable localization in FR images. However, our evaluation on images inpainted with random, non-semantic masks reveals moderate semantic biases.
Lastly, the super-resolution attack exposes the risk that routine GenAI-based image enhancement can erase traces used for forensic analysis.
}
In summary, TGIF2 provides a more comprehensive dataset and evaluation framework that exposes current limitations and guides the development of more effective forensic methods for the detection and localization of realistic, text-guided inpainting manipulations.

The remainder of this paper is organized as follows.
Section~\ref{sec:related} reviews related work in generative media forensics and existing datasets.
Section~\ref{sec:tgif} presents the TGIF2 dataset, including the authentic data sources, inpainting models, workflow, and resulting subsets.
Section~\ref{sec:benchmark} reports our forensic benchmark, covering the performance of existing IFL and SID methods, fine-tuning experiments, the generative quality, and the impact of super-resolution attacks.
Finally, Section~\ref{sec:conclusion} discusses our findings and concludes the paper.

\section{Related Work}\label{sec:related}
This section briefly discusses recent advances in IFL and SID methods, in Section~\ref{sec:related-ifl} and Section~\ref{sec:related-sid}, respectively. Then, Section~\ref{sec:related-datasets} discusses related datasets.

\subsection{Image Forgery Localization}\label{sec:related-ifl}
Image forgery localization aims to identify the regions of an image that have been manipulated. Approaches in this area can be roughly grouped into two categories. Some methods focus on detecting subtle statistical inconsistencies, for example traces of compression~\cite{mareen2022comprint} or sensor noise patterns~\cite{Cozzolino2019Noiseprint}. Others exploit more visible cues, such as unnatural transitions along the boundary between pristine and altered content~\cite{dong2022mvss}. To improve robustness, several works combine multiple forensic signals within a unified framework~\cite{triaridis2024mmfusion, karageorgiou2024fusion, mareen2023harmonizing}. 
Some existing research also considers forgeries created with inpainting techniques. Specialized IFL methods in this setting include detectors that leverage distortions in the Laplacian domain~\cite{li2017localization}, or that learn discriminative features directly from data~\cite{wu2021iid, ijcai2021noisedoesntlie}. These approaches, however, have largely been outperformed by more general methods designed to work across different manipulation types~\cite{liu2022pscc, guillaro2023trufor}. 

Current state-of-the-art models~\cite{liu2022pscc, hu2020span, wu2022ImageForensicsOSN, dong2022mvss, wu2019mantranet, kwon2022catnetv2, guillaro2023trufor} are mostly based on deep neural networks and are trained on datasets covering a wide variety of manipulations, such as splicing, copy–move, or conventional inpainting.
Related to inpainting, PSCC-Net~\cite{liu2022pscc} was partly trained on manipulated data using a conventional inpainting method~\cite{li2020recurrent}, and \new{MantraNet~\cite{wu2019mantranet}} was partly trained on images manipulated with OpenCV's inpainting method. 
More datasets that include inpainted images are discussed in Section~\ref{sec:related-datasets}.
While training on a variety of manipulations makes models broadly applicable, these existing models have not been optimized for the recent case of text-guided image manipulation. 

A complementary line of work has examined laundering attacks, where generative models erase forensic traces while leaving the semantic content unchanged~\cite{cannas2024jpegai}. These results demonstrate that generative compression, or diffusion-based regeneration in general, can serve as effective laundering mechanisms, substantially weakening IFL methods.
The TGIF2 dataset proposed in Section~\ref{sec:tgif} directly addresses these challenges by including fully regenerated manipulated content, thereby enabling systematic evaluation of IFL robustness in these challenging scenarios. Additionally, the impact of generative super resolution is evaluated in the TGIF2 benchmark, in Section~\ref{sec:benchmark-sr}.

\subsection{Synthetic Image Detection}\label{sec:related-sid}
Synthetic image detection methods aim to distinguish between authentic and AI-generated visual content. The field initially developed alongside the rise of Generative Adversarial Network (GAN)-based image synthesis, and many early detectors were trained specifically on GAN outputs~\cite{wang2019cnngenerated, frank2020fredetect, ojha2023univFD, liu2020gramnet, tan2023LGrad, tan2024NPR, zhong2023patchcraft}. With the rapid adoption of (latent) diffusion models (LDM or DMs), such as Stable Diffusion (SD)~\cite{rombach2022sd}, more recent work has focused on this paradigm or explores detectors that can generalize across both GANs and DMs~\cite{corvi2023dimd, wang2023dire, koutlis2024Rine, sha2023defake, karageorgiou2025spai, guillaro2024biasfree}.
As these methods are trained and evaluated on academic datasets, in practice, their performance often decreases when evaluated on images found in the wild. However, their performance can be significantly improved by leveraging in-the-wild content for training~\cite{konstantinidou2025navigating}.

Most approaches rely on the idea that generative models leave behind subtle traces or fingerprints, which can be captured in various feature domains. The type of representation is critical for generalization. For instance, some detectors trained purely on GAN-generated data are still able to identify DM-generated images, highlighting the persistence of shared artifacts across generative families~\cite{ojha2023univFD, tan2024NPR, zhong2023patchcraft}.

Despite this progress, most SID research assumes a binary scenario where an entire image is either authentic or synthetic. This setting overlooks more subtle manipulations, such as text-guided inpainting, where a forged region is blended with pristine content.
In such cases, SID methods produce a global score but cannot localize the edited area.
Moreover, authentic images that are post-processed using generative models without altering their semantics have been shown to trigger SID methods~\cite{mandelli2024laundering, cannas2024jpegai}.

The TGIF2 dataset introduces subsets where manipulated content is confined to local regions but fully regenerated through advanced diffusion-based editing. This allows us to examine not only the limits of SID methods, but also their vulnerability to laundering-style manipulations during generative editing.
Moreover, the impact of generative super resolution on SID methods is evaluated in Section~\ref{sec:benchmark-sr}.

\subsection{\new{Text-Guided Inpainting} Datasets}\label{sec:related-datasets}
A wide range of datasets has been developed for image forensics, covering both IFL and SID, as summarized in several recent surveys~\cite{verdoliva2020media, lin2024detecting}. Many of these resources focus on specific manipulation types such as splicing, copy-move, or deepfakes, while others aim to provide more diverse manipulations.  
For inpainting, earlier datasets such as NIST16~\cite{guan2019mfc} and DEFACTO~\cite{mahfoudi2019defacto} include localized edits, but these were created with non-AI-based inpainting tools and do not involve text guidance.

\paragraph{\new{Early Text-Guided Inpainting Datasets}}
Recently, several datasets with text-guided inpainted images have been released. First, COCO-Glide~\cite{guillaro2023trufor} includes 512 images inpainted with GLIDE~\cite{nichol2021glide}. While valuable as a proof of concept, COCO-Glide is relatively small and limited to low-resolution crops (256$\times$256 px) from MS-COCO~\cite{lin2014coco}.
Additionally, the Autosplice dataset~\cite{jia2023autosplice} contains approximately 3k inpainted spliced images using DALL-E 2.
Most importantly, both datasets do not account for differences between spliced and fully regenerated edits.

\paragraph{\new{Fully Regenerated Inpainted Images}}
TGIF~\cite{mareen2024tgif} presents approximately 75k images manipulated using three text-guided inpainting techniques (SD2, SDXL, and Adobe Firefly) at higher resolutions (up to 1024$\times$1024 px), sourced from approximately 3k images from MS-COCO.
TGIF also differentiates between spliced and fully regenerated images\new{, which was one of the key contributions of the work}. 
After the release of TGIF, several new datasets on text-guided inpainting were proposed.
Most notably, SAGI~\cite{giakoumoglou2025sagi} provides approximately 95k manipulated images sourced from 3 datasets (MS-COCO, RAISE~\cite{dang2015raise}, and OpenImages~\cite{kuznetsova2020openimages}), with resolutions up to 2048$\times$2048 px. SAGI uses advanced pipelines (HD-Painter, Brush-Net, PowerPaint, ControlNet~\cite{zhang2023controlnet}, and Inpaint-Anything) to semantically alter images, based on SD2 (a DM) and LaMa~\cite{suvorov2021lama} (a GAN-based approach). To the best of our knowledge, SAGI is the only other existing dataset (i.e., apart from TGIF) that differentiates between spliced and fully regenerated images.

\paragraph{\new{Other AI-based Inpainting Datasets}}
COinCO~\cite{yang2025coinco} provides approximately 97k manipulated images from COCO, using SD2 and out-of-context prompts.
Furthermore, the GRE dataset~\cite{sun2024gre} presents approximately 228k images inpainted using six generative editing methods and pipelines (LaMa~\cite{suvorov2021lama}, MAT~\cite{li2022mat}, SD2, ControlNet~\cite{zhang2023controlnet}, PaintByExample, and Adobe Photoshop).
\new{The UniAIDet~\cite{zhang2025uniaidet} benchmark includes a dataset of 10k inpainted images, only considering splicing.}
\new{None of these inpainting datasets consider local manipulation using new generative models such as FLUX.1.}
\new{To the best of our knowledge, local manipulation with FLUX.1 was only included in the OpenSDI~\cite{wang2025opensdi} dataset, which additionally includes} SD1.5, SD2.1, SDXL, SD3, to comprise approximately 150k images that are either globally synthesized or locally manipulated.
Note that they do not consider local manipulation with full regeneration.

\paragraph{\new{Going Beyond Object Replacement To Limit Bias}}
\new{All of the above datasets inpaint objects in images, which we hypothesize may introduce a potential bias when using them for training or evaluation. There are few datasets that go beyond object replacement, namely Beyond the Brush (BtB)~\cite{bertazzini2024beyond}, GIM~\cite{chen2025gim}, and BR-Gen~\cite{cai2025brgen}.}
\new{In BtB~\cite{bertazzini2024beyond}, images are inpainted using Fooocus, which internally uses SDXL. Objects are automatically segmented from images, and a vision-language model is used to generate appropriate inpainting prompts (spanning from replacing the object with a new object type, to removing it altogether).}
\new{Next, GIM~\cite{chen2025gim} is a million-scale dataset of inpainted images, either by replacing objects using SD2 and GLIDE or removing them using the Denoising Diffusion Null-Space Model (DDNM)~\cite{wang2022ddnm}.}
Finally, BR-Gen~\cite{cai2025brgen} presents approximately 150k locally forged images, including not only global and main object inpaintings, but also secondary object and background inpaintings, therefore accounting for salient object bias. As generative editing methods, they use LaMa~\cite{suvorov2021lama} and MAT~\cite{li2022mat} as GAN-based approaches and SDXL~\cite{podell2023sdxl}, BrushNet~\cite{ju2024brushnet} and PowerPaint~\cite{zhuang2024powerpaint} as diffusion-based pipelines.
%
Bias is an important challenge in dataset design.
Recent studies~\cite{guillaro2024biasfree, smeu2025circumventing} have shown that forensic models can exploit shortcuts or artifacts that are specific to a dataset rather than to the underlying manipulation process. This not only inflates reported performance but also limits cross-dataset generalization. Designing datasets that minimize such biases is therefore critical for robust training and evaluation.
\new{Hence, in our work, we carefully inspect potential semantic biases, originating from training on object-only replacement.}

\paragraph{\new{Summary of Text-Guided Inpainting Datasets}}
An overview of these datasets with generative inpainted images is given in Table~\ref{tab:related-datasets}, where we focus on spliced vs. fully regenerated images, the used generative models and pipelines, as well as the presence of a subset that goes beyond object replacement.
In summary, no dataset exists that covers all three of the following aspects.
First, \new{only OpenSDI} includes local manipulation with recent generative models such as FLUX~\cite{flux2024}, which has established itself as one of the strongest diffusion-based editors.
Second, \new{only TGIF and SAGI} explicitly differentiate between spliced and fully regenerated images, whereas providing localization in the latter is still an open problem.
Third, few datasets \new{(i.e., only BtB, GIM, and BR-Gen)} consider potential object-based biases and provide non-object-based generative inpainted images.
\new{The lack of a dataset that addresses all these gaps} has motivated the creation and release of the TGIF2 dataset.

\begin{table*}[t]
\centering
\caption{Overview of datasets with generative inpainted images. SP = spliced, FR = fully regenerated.}
\label{tab:related-datasets}
\small
\begin{tabular}{lccP{1.2cm}P{4.6cm}P{1.5cm}}
\toprule
Name & SP & FR & {\# fake \newline images} & Gen.~models \& pipelines & {Beyond \newline obj.~repl.} \\
\midrule
Coco-Glide~\cite{guillaro2023trufor} & \checkmark & - & 512 & GLIDE & \Centering{-} \\
AutoSplice~\cite{jia2023autosplice} & \checkmark & - & 3k & DALL-E2 & - \\
TGIF~\cite{mareen2024tgif} & \checkmark & \checkmark & 75k & SD2, SDXL, Adobe Firefly & - \\
SAGI~\cite{giakoumoglou2025sagi} & \checkmark & \checkmark & 95k & SD2, LaMa, HD-Painter, BrushNet, PowerPaint, ControlNet, Inpaint-Anything & - \\
COinCO~\cite{yang2025coinco} & \checkmark & - & 97k & SD2 & - \\
GRE~\cite{sun2024gre} & \checkmark & - & 228k & LaMa, MAT, SD2, ControlNet, PaintByExample, Photoshop & - \\
\new{UniAIDet~\cite{zhang2025uniaidet}} & \new{\checkmark} & \new{-} & \new{10k} & \new{SD2, SDXL, SD3, DreamShaper, PaintByExample} & \new{-} \\
OpenSDI~\cite{wang2025opensdi} & \checkmark & - & 150k & SD1.5, SD2.1, SDXL, SD3, FLUX.1 & - \\
\new{BtB~\cite{bertazzini2024beyond}} & \new{\checkmark} & \new{-} & \new{22k} & \new{Fooocus (SDXL)} & \new{\checkmark} \\
\new{GIM~\cite{chen2025gim}} & \new{\checkmark} & \new{-} & \new{1.1M} & \new{SD, GLIDE, DDNM} & \new{\checkmark} \\

BR-Gen~\cite{cai2025brgen} & \checkmark & - & 150k & LaMa, MAT, SDXL, BrushNet, PowerPaint & \checkmark \\
\midrule
\textbf{TGIF2}  (ours) & \checkmark & \checkmark & \new{271k} & SD2, SDXL, Adobe Firefly, FLUX.1 schnell/dev/fill-dev & \checkmark \\
\bottomrule
\end{tabular}
\end{table*}

\section{TGIF2 Dataset}\label{sec:tgif}
The TGIF2 dataset is an extension of the original TGIF dataset~\cite{mareen2024tgif}.
In general, compared to existing datasets in Table~\ref{tab:related-datasets}, the TGIF2 dataset is the only dataset that \new{has all three of the following properties:} (1) includes newer generative models (i.e., FLUX.1 models), (2), explicitly differentiates between spliced and fully regenerated images, and (3) goes beyond object replacement (by additionally using random, non-semantic masks).
TGIF2 is available for download at \url{https://github.com/IDLabMedia/tgif-dataset}.

The remainder of this section is organized as follows.
Section~\ref{sec:tgif-source} describes the source of authentic images and their properties.
Section~\ref{sec:tgif-models} details the inpainting models used to generate manipulations.
Section~\ref{sec:tgif-workflow} explains the automated workflow for generating the variety of inpainted images, ending with an overview of the different subsets in Table~\ref{tab:tgif-subsets}.
Finally, Section~\ref{sec:tgif-splits} quantifies the dataset size and splits.

\subsection{Source of Authentic Images}\label{sec:tgif-source}
TGIF2 uses the same source of authentic images as TGIF, namely the MS-COCO dataset~\cite{lin2014coco} (\emph{val2017}), which is licensed under a Creative Commons Attribution 4.0 License. COCO provides images with associated captions, segmentation masks, and object bounding boxes, covering 80 object categories. To comply with restrictions in Adobe Firefly, we excluded two categories (\emph{knife} and, surprisingly, \emph{frisbee}) that were not allowed as prompts in the app.
From each object category, a maximum of 50 images were selected. Images were filtered to ensure both dimensions are at least 512 pixels, while the free Flickr service limited the dimensions to 1024 pixels each. We additionally required object bounding boxes to be smaller than 512$\times$512 px and larger than $64^2$ px in area. This ensures sufficient resolution for high-fidelity inpainting while avoiding excessively small objects that may produce ambiguous manipulations.
These selection and filtering criteria ensure that TGIF2 builds on a diverse yet well-controlled set of authentic images suitable for high-quality and consistent inpainting experiments.

\subsection{Inpainting Models}\label{sec:tgif-models}
TGIF2 includes several inpainting models to cover a wide range of generative capabilities. Most notably, the three FLUX.1 model variants are new in TGIF2, compared to TGIF~\cite{mareen2024tgif}. \new{Note that the images inpainted with PS, SD2 and SDXL are exactly equal in TGIF and TGIF2, with the exception of those with random non-semantic masks (as explained further in Section~\ref{sec:tgif-workflow}).}

\begin{itemize}
\item \textbf{Adobe Firefly (PS)}: commercial model accessed through Photoshop 25.4.0, in early 2024. This produces spliced inpainted regions with an automatically added gradient border.
\item \textbf{Stable Diffusion 2 (SD2)}: a first-generation open-source diffusion model. \new{We used the inpainting-fine-tuned model fine-tuned (\emph{stabilityai/stable-diffusion-2-inpainting}).}
\item \textbf{Stable Diffusion XL (SDXL)}: a second-generation open-source diffusion model that supports higher-resolution outputs (i.e., up to 1024$\times$1024 px) and improved visual quality. \new{We used the inpainting-fine-tuned model (\emph{stabilityai/stable-diffusion-xl-1.0-inpainting-0.1}).}
\item \textbf{FLUX.1 models}: a third-generation open-source inpainting architecture, offering improved attention mechanisms and context-aware generation. Specifically, three FLUX.1 variants are included:
\begin{itemize}
\item \textbf{FLUX.1 [schnell]}: model capable of generating high-quality images in only 1 to 4 diffusion steps. It is trained using latent adversarial diffusion distillation (based on the closed-source FLUX.1 [pro]), hence the optimized speed comes at the cost of a decrease in image quality. 
\item \textbf{FLUX.1 [dev]}: model optimized to balance speed with image quality. Trained using guidance distillation (based on the closed-source FLUX.1 [pro]), making FLUX.1 [dev] more computationally efficient yet producing slightly lower quality images.
\item \textbf{FLUX.1 Fill [dev]}: same architecture as FLUX.1 [dev], but fine-tuned for the inpainting task.
\end{itemize}
\end{itemize}

\new{Note that FLUX.1 [schnell] and FLUX.1 [dev] were trained for full synthetic image generation, in contrast to FLUX.1 Fill [dev] and the SD models which were fine-tuned for the inpainting use case. However, these full-image generation models can be plugged into the inpainting workflow without any adaptations, and perform (surprisingly) well for this use case. In fact, on average, they produce higher quality inpaintings than FLUX.1 Fill [dev] (see Section~\ref{sec:benchmark-quality}).
Using the FLUX.1 [schnell] model is especially practical, as it requires much fewer diffusion steps than FLUX.1 Fill [dev] and hence has significantly lower computational complexity (as further discussed in Section~\ref{sec:tgif-workflow}).
}

\new{It should additionally be noted that FLUX.1 [pro] is available as well, although only through a pay-per-use API. Closed-source models behind APIs do not allow easy reproducibility by the community, and they only provide limited control over their inputs, making them unsuitable for an in-depth forensic analysis. Therefore, we did not include this generative model in our dataset.}

\begin{figure*}[h]
\centering
\subfloat[Original]{\includegraphics[width=0.24\linewidth]{resources/94336/94336_orig.png}
\label{fig:tgif-models-orig}}
\subfloat[Mask (bounding box)]{\includegraphics[width=0.24\linewidth]{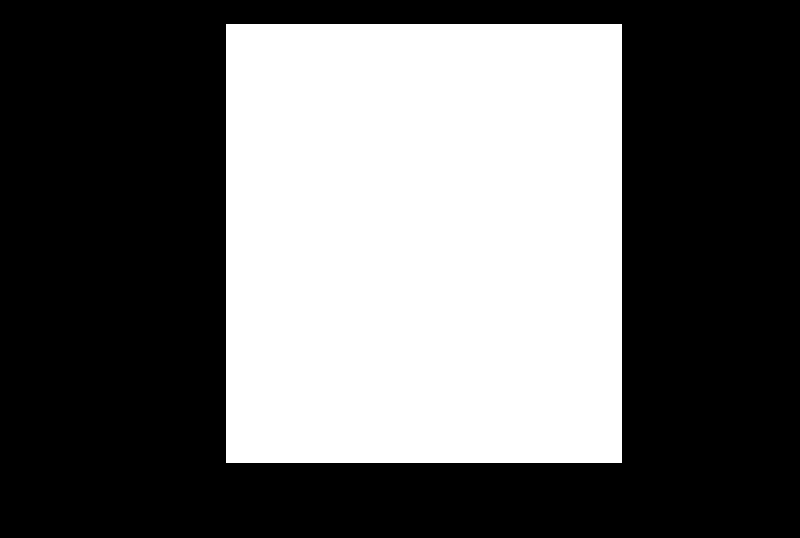}
\label{fig:tgif-models-mask}}
\subfloat[SD2]{\includegraphics[width=0.24\linewidth]{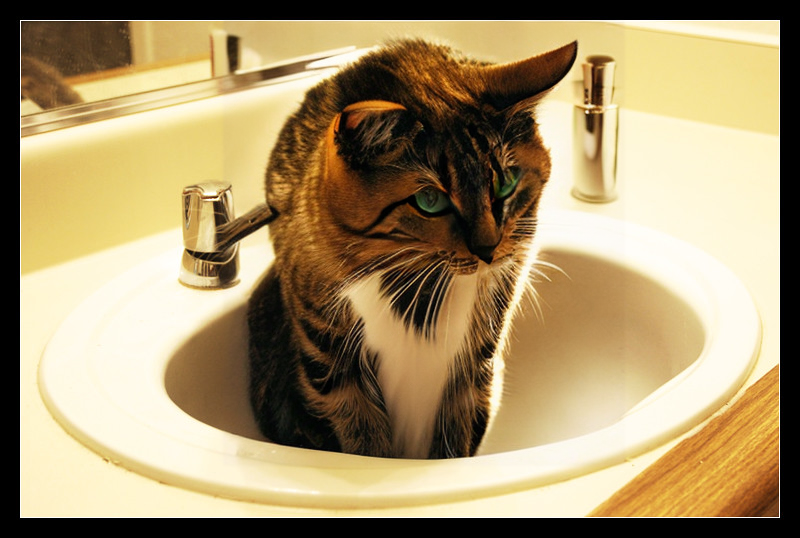}
\label{fig:tgif-models-sd2}}
\subfloat[SDXL]{\includegraphics[width=0.24\linewidth]{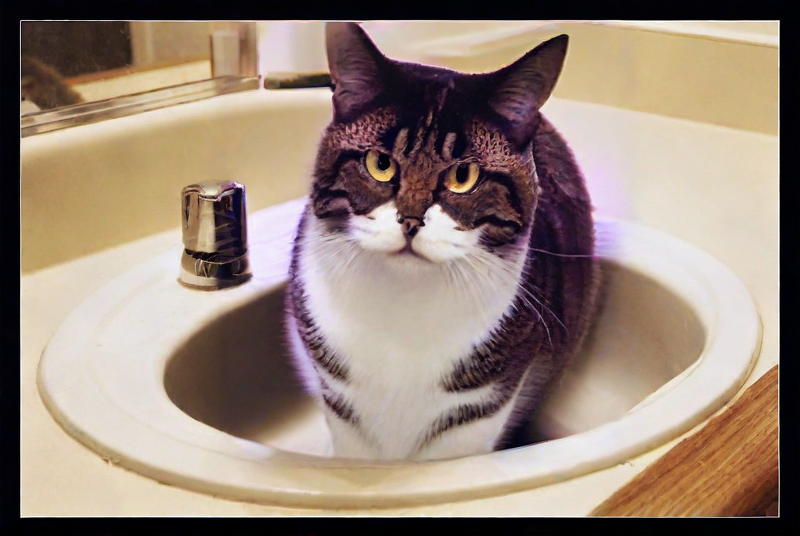}
\label{fig:tgif-models-sdxl}}
\\
\subfloat[PS]{\includegraphics[width=0.24\linewidth]{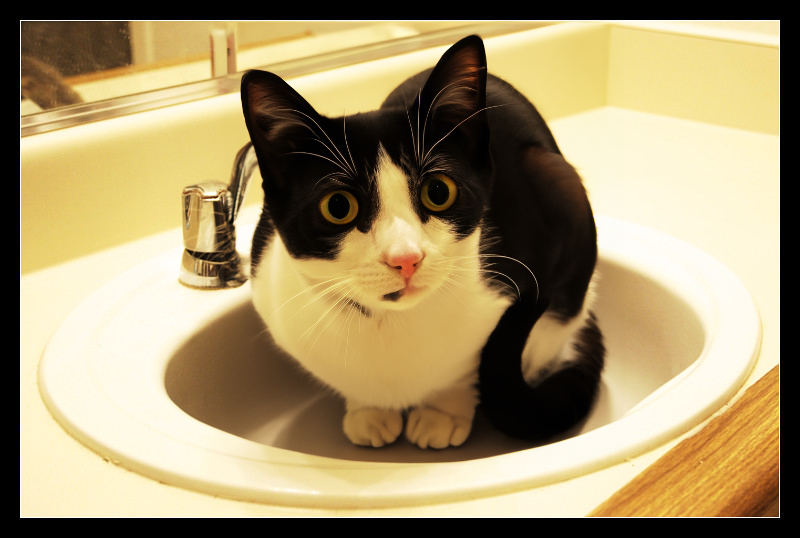}
\label{fig:tgif-models-ps}}
\subfloat[FLUX.1 schnell]{\includegraphics[width=0.24\linewidth]{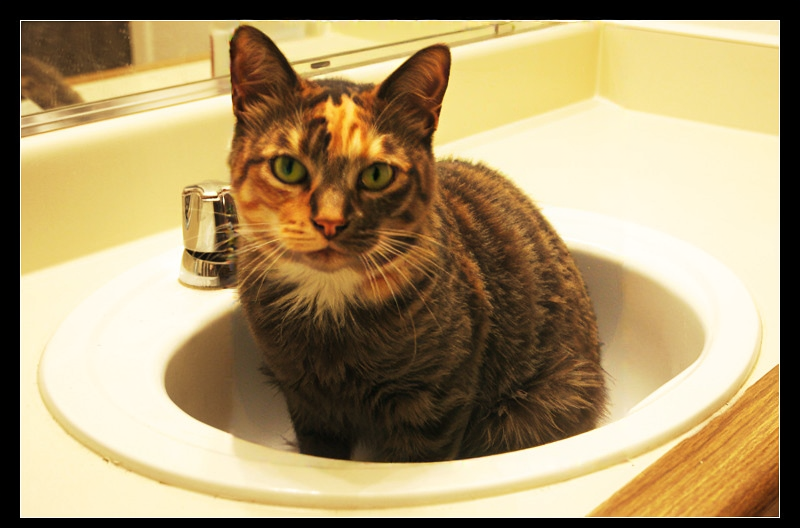}
\label{fig:tgif-models-flux1schnell}}
\subfloat[FLUX.1 dev]{\includegraphics[width=0.24\linewidth]{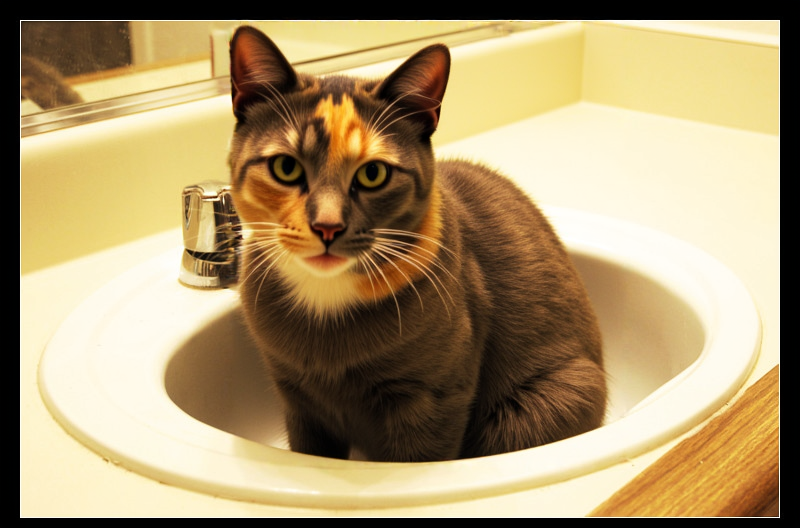}
\label{fig:tgif-models-flux1dev}}
\subfloat[FLUX.1 Fill dev]{\includegraphics[width=0.24\linewidth]{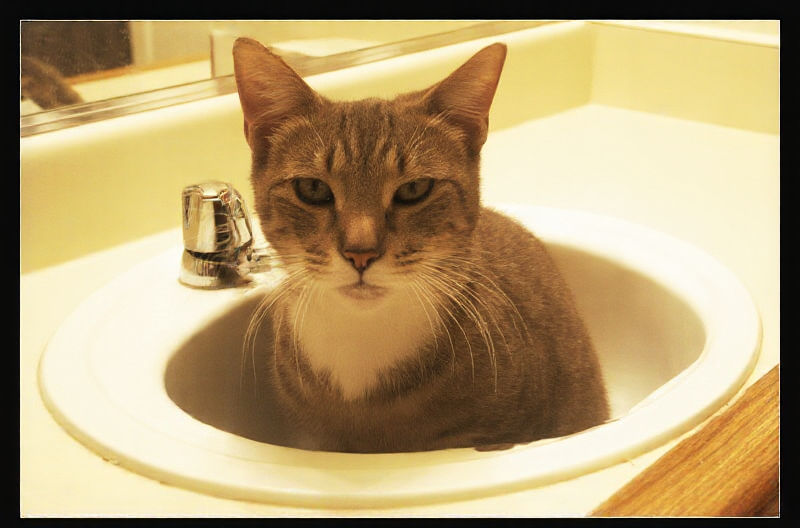}
\label{fig:tgif-models-flux1filldev}}

\caption{Side-by-side comparison of (a) a real image of a cat with (b) the mask used for inpainting, and (c)-(h) 6 inpainted versions using 6 different inpainting models used in TGIF2, respectively.
Note that other masks can be used as well (see Fig.~\ref{fig:tgif-workflow-masks}).
\label{fig:tgif-models}}
\end{figure*}

Fig.~\ref{fig:tgif-models} shows a side-by-side comparison of an example authentic image, the bounding box used as mask, and six inpainted images generated by the six GenAI models, respectively. Note that, in this example, Fig.~\ref{fig:tgif-models-sd2} and~\ref{fig:tgif-models-ps} (inpainted using SD2 and PS, respectively) are spliced, whereas the other inpainted images are fully regenerated. Additionally, a bounding box was used as mask.
Together, these models span commercial and open-source approaches across three generations of diffusion architectures, enabling TGIF2 to capture a broad spectrum of inpainting characteristics.

\subsection{Inpainting Workflow}\label{sec:tgif-workflow}
\begin{figure*}[t]
\centering
\subfloat[Bounding box mask]{\includegraphics[width=0.33\linewidth]{resources/94336/94336_mask_bbox.png}
\label{fig:tgif-workflow-masks-bbox}}
\subfloat[Segmentation mask]{\includegraphics[width=0.33\linewidth]{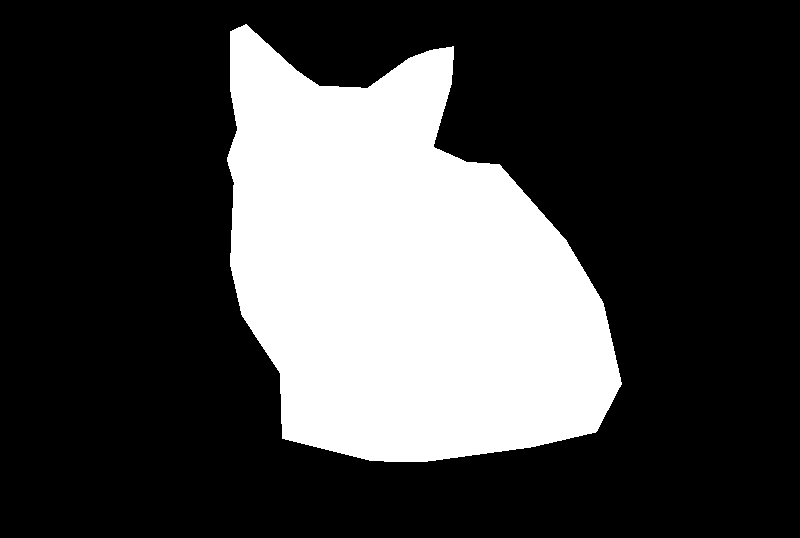}
\label{fig:tgif-workflow-masks-segm}}
\subfloat[\new{Random mask}]{\includegraphics[width=0.33\linewidth]{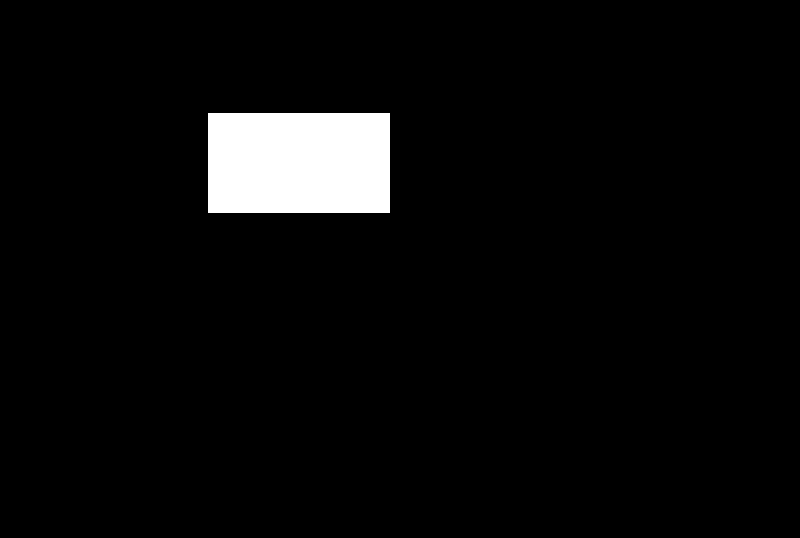}
\label{fig:tgif-workflow-masks-random}}
\\
\subfloat[Inpainted bbox mask]{\includegraphics[width=0.33\linewidth]{resources/94336/94336_mask_bbox.png_ps_mask.png_sd2_0.png}
\label{fig:tgif-workflow-masks-bbox-inp}}
\subfloat[Inpainted segm. mask]{\includegraphics[width=0.33\linewidth]{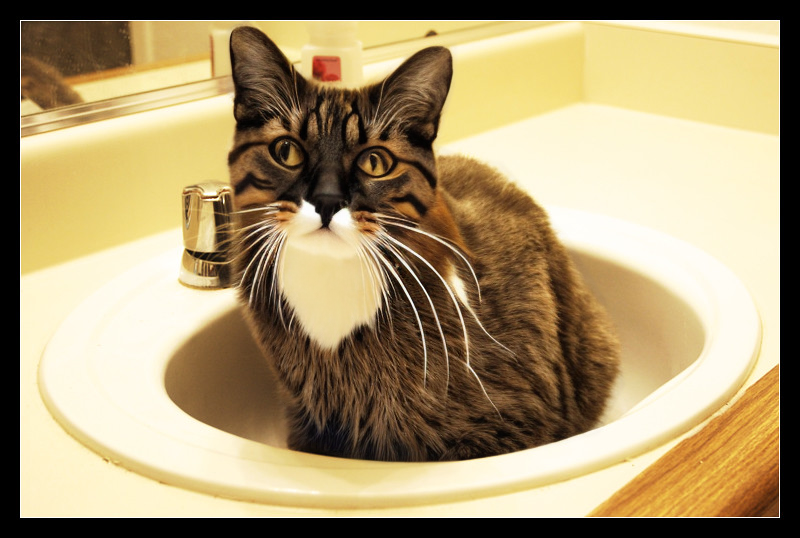}
\label{fig:tgif-workflow-masks-segm-inp}}
\subfloat[\new{Inpainted random mask}]{\includegraphics[width=0.33\linewidth]{resources/94336/94336_mask_random.png_spliced_sd2_2.png}
\label{fig:tgif-workflow-masks-random-inp}}
\caption{The (a)-(c) three types of masks used and (d)-(f) three corresponding inpainted examples, respectively. All inpainted images used Fig.~\ref{fig:tgif-models-orig} as input image.
\label{fig:tgif-workflow-masks}}
\end{figure*}

\new{The inpainting workflow is exactly the same for all inpainting models, with the only difference being the generative model that is called. In particular, }for each authentic image, inpainted variants are automatically generated following these steps:

\begin{itemize}
\item \textbf{Mask Selection}: For each selected image, we use either the object's semantic segmentation mask, the object's bounding box (bbox) mask (both provided by COCO), or a random non-semantic mask.
\new{The random, non-semantic masks are at a random location in the (cropped) image, and with a random rectangular size (min. 64 px and max. 60\% of the image's width and height).}
Examples of these three mask types and corresponding inpainted versions are shown in Fig.~\ref{fig:tgif-workflow-masks}. 

\item \textbf{Prompt Generation}: For SD2, SDXL and FLUX.1, the prompt combines the object category and the image caption (provided by COCO).
During initial experiments, we noticed that including the caption resulted in better results, which was also observed in related work~\cite{xie2023smartbrush, wang2023imagen}.
For Adobe Firefly, only the object category is used, as we noticed that this was sufficient in our initial
experiments. For the random, non-semantic mask, an empty prompt is used.
\item \textbf{Generation Parameters}: Inference steps, guidance scale, and seed were randomly selected per image. For each model and each mask, three variations were generated with three different seeds.
The number of inference steps is between 10 and 50 for the SD and FLUX models, and between 3 and 6 for FLUX.1 [schnell]. The guidance scale is between 1 and 10, except for FLUX.1 [schnell] where it is set to 0 (as it does not support this parameter).
\item \textbf{Splicing vs. Full Regeneration}: We save up to two variants of each inpainted image, namely a fully regenerated (FR) and a spliced (SP) version. The differences between these two are illustrated in Fig.~\ref{fig:intro} and explained below.
\begin{itemize}
\item \emph{Full Regeneration (FR)}: When inpainting using generative models, the entire input image is used to condition the diffusion process. While only the masked area is semantically changed, the non-masked area is subtly altered as well \new{(as visualized in Fig.~\ref{fig:intro})}.
\item \emph{Splicing (SP)}: \new{after performing full regeneration,} the inpainted region is inserted \new{(or \emph{spliced})} back into the original image using the provided mask. This approach preserves the rest of the original image and follows the strategy used in common inpainting pipelines, such as Adobe Firefly and HuggingFace diffusers~\cite{huggingface_diffusers}.
\new{This is also the most commonly used inpainting method in existing datasets (see Section~\ref{sec:related-datasets}).}
\end{itemize}
\item \textbf{\new{Generative Quality} Metrics}: Each image also includes aesthetic quality scores (NIMA~\cite{talebi2018nima},  GIQA~\cite{gu2020giqa}) and image-text-matching (ITM) scores~\cite{li2022blip}, measured on the object's inpainted area.
\new{Additionally, we include the preservation fidelity, calculated using the Peak Signal-to-Noise Ratio (PSNR), Structural SIMilarity index measure (SSIM), and the LPIPS perceptual metric~\cite{zhang2018lpips} between the real and FR image, in the real but fully regenerated areas.}
\end{itemize}

Each of the steps above are applied on each inpainting model, with some exceptions and notes.
First, SD2 is limited to processing images with resolution $512\times512$, which is the resolution of the fully regenerated images in the corresponding subsets. In contrast, for SDXL and the FLUX.1 variants, the resolution is up to 1024p. This only affects the fully regenerated variant, and not the spliced variant.
Second, Adobe Firefly (Photoshop, PS) only provides a spliced variant, but no full regeneration variant.
Third, we do not provide the spliced variant for SDXL, as it discolors the entire image, which is extremely visible when splicing into the original (and a known issue~\cite{diffusers_github_issue}). This discoloration can also be observed when comparing Fig.~\ref{fig:tgif-models-orig} with ~\ref{fig:tgif-models-sdxl}.
\new{Fourth, we do not generate random masks for Adobe Firefly / PS, since it only produces spliced edits, which is the least challenging use case (as discussed in Section~\ref{sec:benchmark-ifl}.
}

Table~\ref{tab:tgif-subsets} gives an overview of the \new{19} resulting TGIF2 subsets.

\begin{table}[!t]
\centering
\caption{Overview of TGIF2 dataset subsets grouped by inpainting model. Columns indicate whether spliced (SP), fully regenerated (FR), and Semantic or Random mask variants are included.}
\setlength\tabcolsep{10pt}%
\label{tab:tgif-subsets}
\begin{tabular}{lK{1cm}K{1cm}|K{1cm}K{1cm}}
\toprule
\multirow{2}{*}{Inpainting Model} & \multicolumn{2}{c|}{Semantic} & \multicolumn{2}{c}{Random} \\
& SP & FR & SP & FR \\
\midrule
SD2                  & \checkmark\new{*} & \checkmark\new{*} & \checkmark & \checkmark \\
SDXL                 & - & \checkmark\new{*} & - & \checkmark \\
Adobe Firefly (PS)   & \checkmark\new{*} & - & - & - \\
FLUX.1 [schnell]     & \checkmark & \checkmark & \new{\checkmark} & \new{\checkmark} \\
FLUX.1 [dev]         & \checkmark & \checkmark & \new{\checkmark} & \new{\checkmark} \\
FLUX.1 Fill [dev]    & \checkmark & \checkmark & \new{\checkmark} & \new{\checkmark} \\
\bottomrule
\multicolumn{5}{c}{\new{* These subsets are already included in the original TGIF dataset.}} \\
\end{tabular}
\end{table}

\subsection{Dataset Statistics and Splits}\label{sec:tgif-splits}
Starting from $3,124$ authentic images, TGIF2 contains $18,744$ manipulated images per SP subset or FR subset, and $9,372$ manipulated images per Random (SP or FR) subset (as only one mask type, i.e., a random rectangular one, is used as opposed to two, i.e., a segmentation mask and bounding box). This amounts to a total of \new{$271,788$} manipulated images in TGIF2, spread over the \new{19} subsets presented in Table~\ref{tab:tgif-subsets} \new{(of which 4 subsets containing 74,976 images are already included in the original TGIF dataset).} Furthermore, the dataset is split approximately by 80/10/10\% into training, validation, and test sets, carefully avoiding category overlap and image leakage between splits.
This makes TGIF2 a reliable resource for training and evaluating forensic models.

\section{Forensic Benchmark}\label{sec:benchmark}
We provide a comprehensive benchmark of image forensic methods on the TGIF2 dataset. 
Compared to the original TGIF benchmark, TGIF2 enables a more diverse and challenging evaluation by introducing new subsets, new fine-tuned models, and new attack scenarios.
Section~\ref{sec:benchmark-setup} describes the evaluation setup employed in the experiments.
Then, in Section~\ref{sec:benchmark-ifl}, we evaluate existing IFL methods on the expanded TGIF2 subsets. 
Section~\ref{sec:benchmark-ifl-fine-tune} investigates whether fine-tuning IFL models on our fully regenerated training sets improves their ability to localize manipulations in FR inpainted images while accounting for potential object biases.
\new{Section~\ref{sec:benchmark-quality} analyzes the generative quality and whether it impacts IFL performance.}
Section~\ref{sec:benchmark-sid} extends the benchmark of SID methods by including the new TGIF2 subsets and recent state-of-the-art SID methods.
Section~\ref{sec:benchmark-sr} evaluates the generative super-resolution attack scenario as a post-processing step.

\subsection{Evaluation Setup}\label{sec:benchmark-setup}
We perform evaluations on our TGIF2 test set, separately considering each of the subsets in Table~\ref{tab:tgif-subsets}. This setup allows us to systematically compare model behavior across different manipulation types and mask strategies. 

For IFL methods, as performance metric, we report the mean pixel-level F1 score over manipulated images, using a threshold of 0.5 to decide whether pixels are real or fake.
\new{We additionally provide the mean Intersection over Union (IoU) results in Appendix~\ref{app:ifl-iou}.}
This provides a calibrated measure of localization accuracy, which is important for practical deployment and human interpretability.
We select several recent high-performing AI-based IFL methods, namely the PSCC-Net~\cite{liu2022pscc}, SPAN~\cite{hu2020span}, ImageForensicsOSN~\cite{wu2022ImageForensicsOSN}, MVSS-Net++~\cite{dong2022mvss}, MantraNet~\cite{wu2019mantranet}, CAT-Net~(v2)~\cite{kwon2022catnetv2}, TruFor~\cite{guillaro2023trufor}, and MMFusion (MMF)~\cite{triaridis2024mmfusion}.
\new{During evaluation, we specifically distinguish among the semantic (Sem) and the random (Rand) subsets.}

For SID methods, we report the area under the ROC curve (AUC) as performance metric. 
We opt for AUC rather than the threshold-specific F1, since we noticed that some methods tend to overpredict the fake class, artificially inflating F1, whereas  the AUC better reflects the trade-off between false positives and false negatives across thresholds.  
We select several AI-based SID methods available in the SIDBench framework~\cite{schinas2024sidbench}, namely CNNDetect~\cite{wang2019cnngenerated}, DIMD~\cite{corvi2023dimd}, Dire~\cite{wang2023dire}, FreqDetect~\cite{frank2020fredetect}, UnivFD~\cite{ojha2023univFD}, Fusing~\cite{ju2022fusing}, GramNet~\cite{liu2020gramnet}, LGrad~\cite{tan2023LGrad}, NPR~\cite{tan2024NPR}, RINE~\cite{koutlis2024Rine}, DeFake~\cite{sha2023defake}, and PatchCraft~\cite{zhong2023patchcraft}.
Additionally, new in TGIF2, we incorporate the recent models SPAI~\cite{karageorgiou2025spai}, SPAI-ITW~\cite{konstantinidou2025navigating}, RINE-ITW~\cite{konstantinidou2025navigating}, and B-Free~\cite{guillaro2024biasfree}.
By combining a diverse set of recent SID approaches with a systematic evaluation protocol, we provide a comprehensive benchmark of current forensic capabilities on generative image inpainting.



\subsection{Image Forgery Localization}\label{sec:benchmark-ifl}
We benchmark recent IFL methods on the TGIF2 dataset. 
Recall that, compared to the original TGIF benchmark, TGIF2 includes several new subsets (recent FLUX.1 models and random non-semantic masks).
Table~\ref{tab:benchmark-ifl-sp} and the top part of Table~\ref{tab:benchmark-ifl-fr} report the F1 values for all evaluated IFL methods on the spliced subsets and on the fully regenerated subsets, respectively.
We highlight F1 values above 0.7 in bold font, chosen empirically as a threshold to allow for quickly grasping which methods demonstrate good detection performance.

For the non-random Sem spliced subsets, some IFL methods (i.e., CAT-Net, TruFor, and MMFusion) show decent performance.
This was already reported in the original TGIF work~\cite{mareen2024tgif}, and is in line with our expectations, as the splicing creates significant differences in forensic traces between forged and pristine regions.
When evaluating the Sem FLUX.1 subsets, we notice that CAT-Net, TruFor, and MMFusion still perform well, although there is a performance drop on the latter.

We also consider inpainted images with random masks, for which the results can be observed in the Rand subsets of Table~\ref{tab:benchmark-ifl-sp}.
\new{Compared to the corresponding Sem subsets, we see a significant drop in performance for MMFusion and TruFor, especially on the FLUX.1 [schnell] and [dev] subsets. In contrast, CAT-Net remains highly effective in detecting the spliced region in the Rand subsets, as well as TruFor and MMFusion on the SD2 subset.}
This suggests that MMFusion and TruFor likely suffer from a detection bias towards object boundaries or semantic cues. The limitations exposed in this subsection should be taken into account when deploying these methods in real-world detection systems.

In contrast to spliced images, the FR setting is much more challenging, as was already observed in the original TGIF benchmark~\cite{mareen2024tgif}. These insights are confirmed in the new FR FLUX.1 subsets, for which the average F1 scores are still low (i.e., well below 0.5).
\new{This is the case for both the Sem and Rand subsets.}
In FR images, the clear boundary between forensic traces in manipulated and authentic regions disappears due to the regenerative process of the entire image during manipulation.

In summary, while some recent state-of-the-art IFL methods perform reliably on spliced manipulations across generative editing models, they struggle on fully regenerated images. 
The random-mask experiments further reveal that some methods, such as MMFusion and TruFor, may rely on semantic or object-related biases rather than imperceptible forensic traces.
Together, these findings highlight the need for developing IFL approaches that generalize beyond semantic object boundaries, and that remain effective in fully regenerated images. To further delve into this issue, Section~\ref{sec:benchmark-ifl-fine-tune} evaluates whether we can fine-tune IFL methods in order to detect FR images.

\begin{table*}[!t]
\begin{center}
\caption{Evaluation of image forgery localization methods on the \textbf{spliced (SP)} subsets of our TGIF2 dataset, for both the semantic (Sem) and random (Rand) versions (F1 score). F1 scores above 0.7 are highlighted in bold.
}
\label{tab:benchmark-ifl-sp}
{
\footnotesize
\setlength\tabcolsep{3pt}%
\begin{tabular}{l cc|c|cc|cc|cc|ccc}
\toprule
\multirow{3}{*}{IFL Method}
 & \multicolumn{2}{c|}{\multirow{2}{*}{SD2}} & \multirow{2}{*}{PS} & \multicolumn{6}{c|}{FLUX.1} & \multicolumn{3}{c}{\multirow{2}{*}{Average}}  \\
 & & & & \multicolumn{2}{c|}{[schnell]} & \multicolumn{2}{c|}{[dev]} & \multicolumn{2}{c|}{Fill [dev]} \\
 & Sem & \new{Rand} & Sem &
 \new{Sem} & \new{Rand} & \new{Sem} & \new{Rand} & \new{Sem} & \new{Rand} & \new{Sem} & \new{Rand} & \new{All} \\
\midrule
PSCC-Net~\cite{liu2022pscc} & 0.15 & 0.17 & 0.39 & 0.03 & 0.09 & 0.02 & 0.05 & 0.08 & 0.21 & 0.13 & 0.13 & 0.13 \\
SPAN~\cite{hu2020span}  & 0.00 & 0.00 & 0.00 & 0.00 & 0.00 & 0.0& 0.00 & 0.00 & 0.01 & 0.00 & 0.00 & 0.00 \\
ImageForensicsOSN~\cite{wu2022ImageForensicsOSN} & 0.23 & 0.09 & 0.36 & 0.31 & 0.11 & 0.23 & 0.08 & 0.12 & 0.10 & 0.25 & 0.09 & 0.18 \\
MVSS-Net++~\cite{dong2022mvss} & 0.07 & 0.03 & 0.08 & 0.15 & 0.06 & 0.11 & 0.03 & 0.09 & 0.09 & 0.10 & 0.05 & 0.08 \\
Mantranet~\cite{wu2019mantranet} & 0.15 & 0.15 & 0.56 & 0.06 &0.02 & 0.04 & 0.02 & 0.06 & 0.04 & 0.17 & 0.06 & 0.12 \\
CAT-Net~\cite{kwon2022catnetv2} & \textbf{0.89} & \textbf{0.92} & \textbf{0.86} & \textbf{0.88} & \textbf{0.93} & \textbf{0.87} & \textbf{0.92} & \textbf{0.86} & \textbf{0.93} & \textbf{0.87} & \textbf{0.92} & \textbf{0.89} \\
TruFor~\cite{guillaro2023trufor} & \textbf{0.83} & \textbf{0.87} & \textbf{0.79} & \textbf{0.79} & 0.69 & \textbf{0.73} & 0.58 & \textbf{0.74} & \textbf{0.72} & \textbf{0.77} & \textbf{0.72} & \textbf{0.75} \\
MMFusion~\cite{triaridis2024mmfusion} & \textbf{0.73} & \textbf{0.74} & \textbf{0.72} & \textbf{0.70} & 0.59 & 0.59 & 0.45 & 0.63 & 0.59 & 0.68 & 0.59 & 0.64 \\
\bottomrule
\end{tabular}
}
\end{center}
\end{table*}

\begin{table*}[!t]
\begin{center}
\caption{Evaluation of image forgery localization methods on the \textbf{fully regenerated (FR)} subsets of our TGIF2 dataset, for both the semantic (sem) and random (rand) versions (F1 score).
The last rows show the IFL methods fine-tuned on the FR subsets.
F1 scores above 0.7 are highlighted in bold.
}
\label{tab:benchmark-ifl-fr}
{
\footnotesize
\setlength\tabcolsep{2pt}%
\begin{tabular}{l cc|cc|cc|cc|cc|ccc}
\toprule
\multirow{3}{*}{IFL Method}
  & \multicolumn{2}{c|}{\multirow{2}{*}{SD2}} & \multicolumn{2}{c|}{\multirow{2}{*}{SDXL}} & \multicolumn{6}{c|}{FLUX.1} & \multicolumn{3}{c}{\multirow{2}{*}{Average}}\\
 & & & & & \multicolumn{2}{c|}{[schnell]} & \multicolumn{2}{c|}{[dev]} & \multicolumn{2}{c|}{Fill [dev]} & & \\
 & Sem & \new{Rand} & Sem & \new{Rand} &
 \new{Sem} & \new{Rand} & \new{Sem} & \new{Rand} & \new{Sem} & \new{Rand} & \new{Sem} & \new{Rand} & \new{All} \\
\midrule
PSCC-Net~\cite{liu2022pscc} %
& 0.05 & 0.04 & 0.05 & 0.09 & 0.02 & 0.02 & 0.02 & 0.02 & 0.03 & 0.05 & 0.03 & 0.04 & 0.04 \\
SPAN~\cite{hu2020span} %
 & 0.00 & 0.00 & 0.00 & 0.00 & 0.00 & 0.00 & 0.00 & 0.00 & 0.00 & 0.00 & 0.00 & 0.00 & 0.00 \\
ImageForensicsOSN~\cite{wu2022ImageForensicsOSN} %
 & 0.20 & 0.09 & 0.18 & 0.09 & 0.30 & 0.08 & 0.23 & 0.07 & 0.09 & 0.05 & 0.20 & 0.08 & 0.13 \\
MVSS-Net++~\cite{dong2022mvss} %
 & 0.06 & 0.03 & 0.09 & 0.05 & 0.15 & 0.03 & 0.12 & 0.03 & 0.07 & 0.04 & 0.10 & 0.04 & 0.07 \\
Mantranet~\cite{wu2019mantranet} %
 & 0.03 & 0.01 & 0.05 & 0.06 & 0.05 & 0.02 & 0.04 & 0.01 & 0.02 & 0.01 & 0.04 & 0.02 & 0.03 \\
CAT-Net~\cite{kwon2022catnetv2} %
 & 0.04 & 0.01 & 0.03 & 0.00 & 0.05 & 0.01 & 0.04 & 0.01 & 0.03 & 0.02 & 0.04 & 0.01 & 0.02 \\
TruFor~\cite{guillaro2023trufor} %
 & 0.19 & 0.07 & 0.18 & 0.09 & 0.33 & 0.09 & 0.25 & 0.06 & 0.16 & 0.08 & 0.22 & 0.08 & 0.14 \\
MMFusion~\cite{triaridis2024mmfusion} %
 & 0.15 & 0.05 & 0.19 & 0.08 & 0.34 & 0.08 & 0.20 & 0.04 & 0.12 & 0.06 & 0.20 & 0.06 & 0.13 \\
 \midrule
TruFor -- fine-tuned %
 & 0.49 & 0.39 & 0.68 & \textbf{0.83} & \textbf{0.87} & \textbf{0.94} & \textbf{0.82} & \textbf{0.88} & \textbf{0.76} & \textbf{0.85}  & \textbf{0.72} & \textbf{0.78} & \textbf{0.75} \\
MMFusion -- fine-tuned %
 & 0.48 & 0.39 & 0.66 & \textbf{0.83} & \textbf{0.86} & \textbf{0.94} & \textbf{0.81} & \textbf{0.87} & \textbf{0.75} & \textbf{0.84}  & \textbf{0.71} & \textbf{0.77} & \textbf{0.74}\\
\bottomrule
\end{tabular}
}
\end{center}
\end{table*}

\newpage

\subsection{Fine-tuning IFL Approaches on Fully Regenerated Images}\label{sec:benchmark-ifl-fine-tune}

\begin{table*}[!t]
\begin{center}
\caption{\new{Evaluation of \textbf{individually fine-tuned} image forgery localization methods on the \textbf{fully regenerated (FR)} subsets of our TGIF2 dataset, for both the semantic (Sem) and random (Rand) versions (F1 score).
F1 scores above 0.7 are highlighted in a bold font.
We notice that in-domain (i.e., same generative method) performance is relatively good, but generalization to out-of-domain (i.e., different generative method) performance is bad.
Interestingly, when fine-tuning on only Sem masks, the evaluation on Rand subsets is lower than the corresponding Sem subsets, suggesting a potentially introduced bias towards semantics.
}
}
\label{tab:benchmark-ifl-fine-tune}
{
\footnotesize
\setlength\tabcolsep{1.5pt}%
\begin{tabular}{lll cc|cc|cc|cc|cc|ccc}
\toprule
\multirow{3}{*}{IFL Method} & \multicolumn{2}{l}{\multirow{3}{*}{Fine-tune set}} & 
    \multicolumn{2}{c|}{\multirow{2}{*}{SD2}} & \multicolumn{2}{c|}{\multirow{2}{*}{SDXL}} & \multicolumn{6}{c|}{FLUX.1} & \multicolumn{3}{c}{\multirow{2}{*}{Average}}\\
 & & & & & & & \multicolumn{2}{c|}{[schnell]} & \multicolumn{2}{c|}{[dev]} & \multicolumn{2}{c|}{Fill [dev]} \\
 & & & Sem & Rand & Sem & Rand &
 Sem & Rand & Sem & Rand & Sem & Rand & Sem & Rand & All \\
 \midrule

\multirow{9}{*}{TruFor}%
& \multirow{3}{*}{All} & Sem+Rand & 0.49 & 0.39 & 0.68 & \textbf{0.83} & \textbf{0.87} & \textbf{0.94} & \textbf{0.82} & \textbf{0.88} & \textbf{0.76} & \textbf{0.85} & \textbf{0.72} & \textbf{0.78} & \textbf{0.75} \\
   &  & Sem %
& 0.50 & 0.23 & 0.61 & 0.62 & \textbf{0.80} & 0.62 & 0.69 & 0.46 & 0.65 & 0.59 & 0.55 & \textbf{0.76} & 0.65 \\
&  & Rand & 0.31 & 0.40 & 0.50 & \textbf{0.80} & 0.69 & \textbf{0.92} & 0.62 & \textbf{0.85} & 0.63 & \textbf{0.82} & 0.55 & \textbf{0.76} & 0.58 \\[6pt]
    
 
  & \multirow{2}{*}{SD2} & Sem %
 & 0.49 & 0.22 & 0.29 & 0.17 & 0.40 & 0.16 & 0.31 & 0.08 & 0.21 & 0.10 & 0.34 & 0.14 & 0.24 \\
   &  & Rand %
 & 0.31 & 0.40 & 0.17 & 0.22 & 0.12 & 0.10 & 0.09 & 0.06 & 0.12 & 0.12 & 0.16 & 0.18 & 0.17 \\
    & \multirow{2}{*}{SDXL} & Sem %
 & 0.09 & 0.01 & 0.66 & \textbf{0.71} & 0.19 & 0.02 & 0.08 & 0.01 & 0.06 & 0.02 & 0.22 & 0.15 & 0.19 \\
    &  & Rand %
 & 0.01 & 0.00 & 0.50 & \textbf{0.82} & 0.02 & 0.00 & 0.01 & 0.01 & 0.03 & 0.03 & 0.11 & 0.17 & 0.14 \\
     & \multirow{2}{*}{Flux} & Sem %
 & 0.10 & 0.02 & 0.20 & 0.05 & \textbf{0.90} & \textbf{0.90} & \textbf{0.86} & \textbf{0.81} & \textbf{0.78} & \textbf{0.79} & 0.57 & 0.51 & 0.54 \\
    &  & Rand %
 & 0.02 & 0.02 & 0.05 & 0.06 & \textbf{0.73} & \textbf{0.95} & 0.68 & \textbf{0.91} & 0.68 & \textbf{0.84} & 0.43 & 0.56 & 0.49 \\
 
 \midrule
 
  \multirow{9}{*}{MMFusion}%
  & \multirow{3}{*}{All} & Sem+Rand %
 & 0.48 & 0.39 & 0.66 & \textbf{0.83} & \textbf{0.86} & \textbf{0.94} & \textbf{0.81} & \textbf{0.87} & \textbf{0.75} & \textbf{0.84} & \textbf{0.71} & \textbf{0.77} & \textbf{0.74} \\
  &  & Sem %
 & 0.47 & 0.17 & 0.58 & 0.61 & \textbf{0.80} & 0.58 & \textbf{0.72} & 0.50 & 0.69 & 0.67 & 0.65 & 0.50 & 0.63 \\
 &  & Rand %
 & 0.26 & 0.40 & 0.48 & \textbf{0.80} & 0.62 & \textbf{0.91} & 0.58 & \textbf{0.83} & 0.62 & \textbf{0.81} & 0.51 &\textbf{ 0.75} & 0.58 \\[6pt]

  & \multirow{2}{*}{SD2} & Sem %
 & 0.52 & 0.22 & 0.28 & 0.19 & 0.35 & 0.15 & 0.27 & 0.09 & 0.18 & 0.09 & 0.32 & 0.15 & 0.23 \\
   &  & Rand %
 & 0.30 & 0.34 & 0.18 & 0.19 & 0.08 & 0.05 & 0.07 & 0.03 & 0.09 & 0.08 & 0.14 & 0.14 & 0.14 \\
    & \multirow{2}{*}{SDXL} & Sem %
 & 0.08 & 0.01 & 0.66 & \textbf{0.70} & 0.12 & 0.01 & 0.06 & 0.01 & 0.05 & 0.02 & 0.19 & 0.15 & 0.17 \\
    &  & Rand %
 & 0.01 & 0.01 & 0.50 & \textbf{0.80} & 0.03 & 0.00 & 0.02 & 0.01 & 0.04 & 0.04 & 0.12 & 0.17 & 0.15 \\
     & \multirow{2}{*}{Flux} & Sem %
 & 0.10 & 0.02 & 0.20 & 0.05 & \textbf{0.89} & \textbf{0.87} & \textbf{0.85} & \textbf{0.76} & \textbf{0.78} & \textbf{0.80} & 0.56 & 0.50 & 0.53 \\
    &  & Rand %
& 0.03 & 0.02 & 0.06 & 0.07 & \textbf{0.73} & \textbf{0.95} & 0.69 & \textbf{0.90} & 0.69 & \textbf{0.84} & 0.44 & 0.56 & 0.50 \\
 
\bottomrule
\end{tabular}
}
\end{center}
\end{table*}

To address the poor performance of existing IFL methods on FR images (see Section~\ref{sec:benchmark-ifl} and Table~\ref{tab:benchmark-ifl-fr}), we fine-tune selected IFL models on the TGIF2 FR training subsets (i.e., SD2, SDXL, Flux.1 [schnell], Flux.1 [dev], and Flux.1 Fill [dev], both with semantic and random masks).
\new{We do this for each subset individually (but combining the three FLUX.1 subsets into one main FLUX.1 subset), as well as for all subsets collectively. In the collective fine-tuning, we sample an equal number of images from SD2, SDXL and FLUX.1 to avoid a bias to FLUX.1, since it has three times more images than SD2 or SDXL.}
As base models, we select TruFor and MMFusion, i.e., two of the IFL models with the best performance on spliced and fully regenerated images. CAT-Net was excluded from fine-tuning experiments due to its significantly more complex training process.
We followed the training setups specified in the original papers for both models.
That is, both models follow a two-stage training procedure. In the first stage, the model is trained on the forgery localization task for 100 epochs while in the second stage, part of the model is frozen and is trained on the forgery detection task for another 100 epochs. Then, the checkpoint with the best validation loss is chosen as the final checkpoint. We maintain the hyperparameters and other training configurations specified in the original papers.
This fine-tuning approach allows us to adapt high-performing IFL models to better localize manipulations in fully regenerated images, while remaining consistent with the original training protocols.

\new{
Table~\ref{tab:benchmark-ifl-fine-tune} reports the F1 scores for the fine-tuned models.
In short, we observe that fine-tuning significantly improves the performance. However, performance gains are substantially larger when training and testing on semantically aligned masks than when evaluated on random-mask FR images. This is discussed more elaborately in the remainder of this section.
}

\begin{figure*}[t]
\centering
\subfloat[Inpainted (SD2-FR-Sem)]{\includegraphics[width=0.33\linewidth]{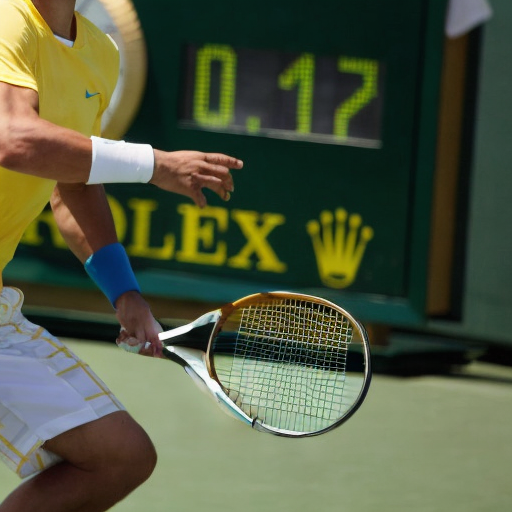}
\label{fig:benchmark-ifl-fine-tune-example-sd2-inpainted}}
\subfloat[Ground-truth mask]{\includegraphics[width=0.33\linewidth]{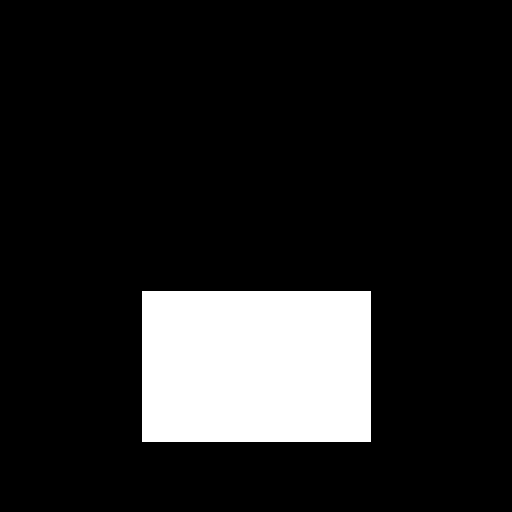}
\label{fig:benchmark-ifl-fine-tune-example-sd2-mask}}
\subfloat[SD2-FR-Sem-finetuned TruFor]{\includegraphics[width=0.33\linewidth]{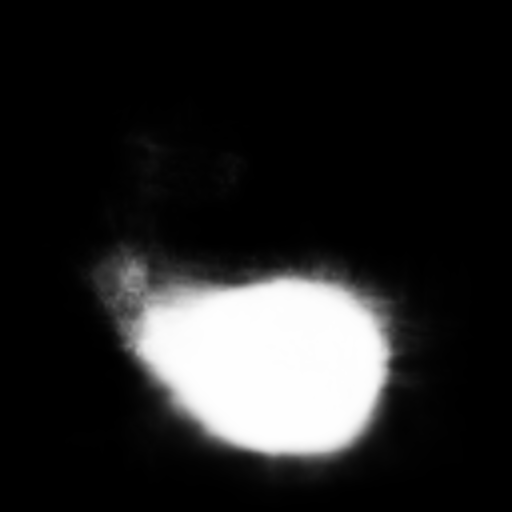}
\label{fig:benchmark-ifl-fine-tune-example-sd2-heatmap}}

\subfloat[Inpainted (SD2-FR-Rand)]{\includegraphics[width=0.33\linewidth]{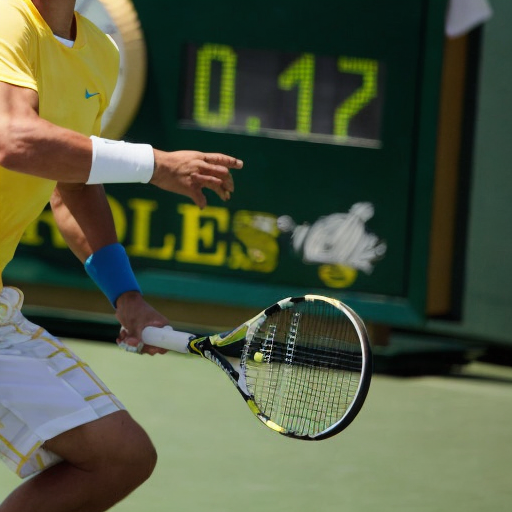}
\label{fig:benchmark-ifl-fine-tune-example-sd2-random-inpainted}}
\subfloat[Ground-truth mask]{\includegraphics[width=0.33\linewidth]{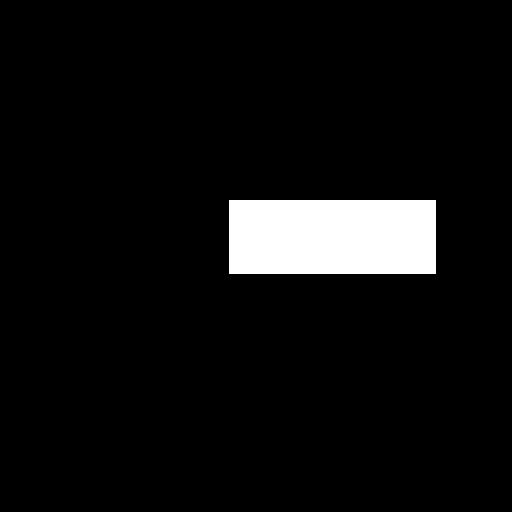}
\label{fig:benchmark-ifl-fine-tune-example-sd2-random-mask}}
\subfloat[SD2-FR-Sem-finetuned TruFor]{\includegraphics[width=0.33\linewidth]{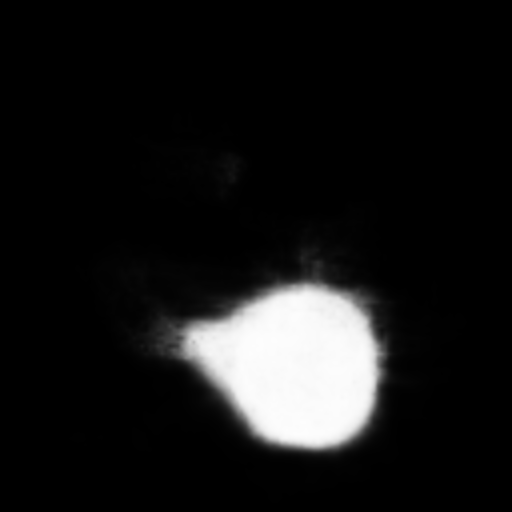}
\label{fig:benchmark-ifl-fine-tune-example-sd2-random-heatmap}}

\caption{\new{An example of a fully regenerated inpainted image with SD2, using a bounding box of the tennis racket as mask (a, SD2 Sem), or a random mask in the image (d, SD2 Rand), along with the corresponding ground-truth masks (b, e) and the fine-tuned TruFor detection results (c, f). The fine-tuned TruFor was trained on the SD2-FR-Sem training set. The inpainted tennis racket in the SD2-FR-Sem test image is detected relatively well by the fine-tuned model. However, the inpainted random box  of the SD2-FR-Rand subset is \emph{not} detected; instead, the tennis racket is wrongly detected -- suggesting that the fine-tuned model is be biased towards semantics or salient objects. Note that the inpainted random box, in fact, made notable changes to the sign in the background, and hence should be detectable by a better detection model.}
\label{fig:benchmark-ifl-fine-tune-example}}
\end{figure*}

\new{
First, we consider the finetuned models on all semantic datasets. The performance on the Sem datasets is relatively high, but is significantly lower for the Random datasets (with the exception of SDXL). We demonstrate this in
Fig.~\ref{fig:benchmark-ifl-fine-tune-example}, which shows an image inpainted (and fully regenerated) using a semantic mask as well as with a random mask, along with the corresponding ground-truth masks and localization results from TruFor finetuned on the SD2-FR-Sem subset. The fine-tuned model is able to localize the semantic object (i.e., the tennis racket), but not the random mask in the background.
Note that the inpainted random box, in fact, made notable changes to the sign in the background, and hence should be detectable by a better detection model
Additionally, note that the object categories do not overlap between the training and test set, i.e., no images of inpainted tennis rackets are seen in the fine-tuned training set.
This indicates that training exclusively on semantic datasets can induce a bias toward salient or semantically meaningful regions.
}

\new{
Second, including the Rand subsets during fine-tuning mitigates the semantic bias observed in the previous setting.
In fact, additionally including random masks during finetuning improves performance on the random subsets without degrading — and in several cases even improving — performance on the semantic subsets.
We select the fine-tuned models on All Sem+Rand subsets as the main one, and also include these results in the bottom rows of Table~\ref{tab:benchmark-ifl-fr}.
}

\new{
Third, beyond mask-related effects, we also observe a discrepancy in IFL performance between SD2, SDXL, and the FLUX.1 family. That is, the IFL performance of FLUX.1 is significantly higher than SD2.
This is especially notable, since we notice contrasting behavior when evaluating the IFL performance on spliced images in Table~\ref{tab:benchmark-ifl-sp}.
Note that we do not only observe this when training on All subsets, but also when training on each subset individually. 
We hypothesized that this discrepancy could be explained by their difference in generative quality, although find no supporting evidence of this, in Section~\ref{sec:benchmark-quality}.
It remains to be investigated, in future work, why FLUX.1 fully regenerated images are easier to detect than SD, after fine-tuning.
}

\new{
Fourth and finally, the models fine-tuned on individual subsets expose that performance on other, out-of-domain generative models is substantially lower than on the in-domain subset that they were trained on.
Note that these methods were originally designed for splicing and copy-move detection.
Hence, an improved training strategies alone may not suffice. Instead, more robust localization mechanisms may be required for localization in fully regenerated images with better generalization.
}


\new{
In summary, this section demonstrated that inpainted areas can be detected in fully regenerated images, which is in line with similar observations in related work that trained or fine-tuned models perform well on FR image data~\cite{smeu2025declip, giakoumoglou2025sagi}. However, fine-tuning exclusively on semantic subsets can induce object-level bias, underscoring the importance of incorporating non-semantic FR data during both training and evaluation.
}

\subsection{\new{Generative Quality vs. IFL Performance}}
\label{sec:benchmark-quality}
\new{
This subsection investigates whether generative quality is predictive of forensic localization performance. Intuitively, higher-quality or more realistic generations might be expected to suppress forensic traces and therefore reduce IFL performance. Conversely, systematic artifacts in lower-quality generations may facilitate detection. We therefore analyze whether aesthetic quality, image–text alignment, and preservation fidelity correlate with IFL performance across subsets and on a per-image level.
We additionally examine whether such quality differences help explain the performance discrepancy between SD and FLUX in fine-tuned FR settings (as observed in Section~\ref{sec:benchmark-ifl-fine-tune}).
}

\new{
As mentioned at the end of Section~\ref{sec:tgif-workflow}, we measure the generative quality metrics in three ways.
First, we measure the aesthetic quality using the NIMA~\cite{talebi2018nima} and GIQA~\cite{gu2020giqa} metrics on the inpainted area.
Second, we measure the image-text matching performance~\cite{li2022blip} on the inpainted area.
Third, we measure the preservation fidelity in real but fully regenerated areas using PSNR, SSIM, and LPIPS.
Finally, we relate these generative quality metrics to the IFL performance in two ways: (1) analyzing trends across the subsets, and (2) we calculate the Spearman's correlation coefficients between per-image F1 scores and each generative quality metric.
We have done this analysis for the TruFor and MMFusion IFL models, namely for the original variants on the SP subsets, and the fine-tuned variants (on All Sem+Rand) on the FR subsets.
Note that we have not done this for the original IFL models evaluated on the FR subsets, as their detection capability is too low in this case (as discussed in Section~\ref{sec:benchmark-ifl}).}

\new{Table~\ref{tab:benchmark-quality-subset} shows the average generative quality metrics for each semantic subset of TGIF2.
For the reader's convenience, we again include the IFL performance of TruFor and MMFusion (as presented before in Table~\ref{tab:benchmark-ifl-sp} and Table~\ref{tab:benchmark-ifl-fr}).
Additionally, Table~\ref{tab:benchmark-quality-corr} shows the correlation values between the generative metrics and the IFL performance, calculated on all semantic subsets simultaneously.
Recall, as also observed previously in Section~\ref{sec:benchmark-ifl} and Section~\ref{sec:benchmark-ifl-fine-tune}, that SD2 has the best and FLUX.1 [dev] has the worst IFL (SP) performance.
In contrast, the IFL (FR) performance is best for FLUX.1 [schnell], and significantly worse for SD2.
Note that, for the preservation fidelity, it only makes sense to relate it with IFL (FR) but not to IFL (SP) performance.
In the following, we inspect whether these performance discrepancies can be explained by any of the generative quality metrics.
}

\new{
Concerning aesthetic quality, in Table~\ref{tab:benchmark-quality-corr}, on a per-subset level, there appears to be no relation with IFL (SP) nor IFL (FR) performance.
In contrast, on a per-image level, in Table~\ref{tab:benchmark-quality-corr}, we notice there is a weak to moderate correlation (i.e., correlation values around 0.3 for NIMA and 0.4 for GIQA) between the aesthetic quality and IFL performance (both for SP and FR).
Interestingly, higher aesthetic scores are weakly associated with better IFL performance at the per-image level, contradicting the intuition that visually appealing generations should be harder to detect.
However, since we do not observe this relation on a per-subset level, and hence may not serve as an explanation as to why FLUX.1 performs significantly better than SD2 in IFL (FR).
}


\new{
For ITM, in Table~\ref{tab:benchmark-quality-subset}, on a per-subset level, we do not observe a relation with the IFL (SP) nor IFL (FR) performance.
On a per-image level, in Table~\ref{tab:benchmark-quality-corr}, we see a negligible to weak negative correlation for ITM (i.e., approx. -0.3 for IFL (SP) and 0 for IFL (FR)), and a negligible to weak positive correlation for the ITM cosine similarity (i.e., approx. 0.1 for IFL (SP) and 0.2 for IFL (FR)).
Hence, we conclude that there is no clear relationship between ITM and IFL performance.}

\new{
For the preservation fidelity, in Table~\ref{tab:benchmark-quality-subset}, on a per-subset level, we observe no correlation with IFL (FR).
On a per-image level, in Table~\ref{tab:benchmark-quality-corr}, we observe only a weak positive correlation (i.e., approx. 0.2 -- 0.3) with IFL (FR).
Hence, we conclude that there is no notable relationship between preservation fidelity and IFL performance.
}

\begin{table*}[!t]
\begin{center}
\caption{\new{Average generative quality per (semantic) subset.
In addition, the average detection performance is given as well (same as in Table~\ref{tab:benchmark-ifl-fr}).
For each metric, we highlight the best performing subset in a bold font, and the worst performing subset in an italic font.
}
}
\label{tab:benchmark-quality-subset}
{
\footnotesize
\setlength\tabcolsep{4pt}%
\begin{tabular}{lll cccccc}
\toprule
\multicolumn{2}{l}{\multirow{3}{*}{Metric Class}} & \multirow{3}{*}{Metric} & 
  \multirow{2}{*}{SD2} & \multirow{2}{*}{PS}  & \multirow{2}{*}{SDXL} & \multicolumn{3}{c}{FLUX.1} \\
 & & & & & & [schnell] & [dev] & Fill [dev] \\
\midrule
\multicolumn{2}{l}{\multirow{2}{*}{Aesthetic Quality}}
& NIMA mean$\uparrow$ & 5.05 & 5.16 & \textit{4.88} & 5.16 & \textbf{5.28} & 5.21 \\
& & GIQA$\uparrow$ & -399 & -373 & -423 & -383 & \textbf{-371} & \emph{-424} \\[6pt]
\multicolumn{2}{l}{\multirow{2}{*}{Image-Text Matching}}
& ITM$\uparrow$ & 0.28 & \textbf{0.32} & 0.26 & 0.31 & 0.30 & \textit{0.25} \\
& & cos$\uparrow$ & 0.37 & \textbf{0.38} & \textit{0.34} & \textbf{0.38} & \textbf{0.38} & 0.36 \\[6pt]
\multicolumn{2}{l}{\multirow{3}{*}{Preservation Fidelity (FR only)}}
& PSNR$\uparrow$ & 27.1 & N/A & \textit{18.1} & \textbf{32.5} & \textbf{32.5} & 28.5 \\
& & SSIM$\uparrow$ & 0.79 & N/A & \textit{0.69} & \textbf{0.92} & \textbf{0.92} & 0.86 \\
& & LPIPS$\downarrow$ & 0.04 & N/A & \textit{0.17} & 0.02 & \textbf{0.01} & 0.06 \\
\midrule
\multirow{2}{*}{IFL (SP)}
& TruFor -- original & F1 score$\uparrow$ & \textbf{0.83} & 0.79 & N/A & 0.79 & \textit{0.73} & 0.74 \\
& MMFusion -- original & F1 score$\uparrow$ & \textbf{0.73} & 0.72 & N/A & 0.70 & \textit{0.59} & 0.63 \\[6pt]
\multirow{2}{*}{IFL (FR)}
& TruFor -- fine-tuned & F1 score$\uparrow$ & \textit{0.49} & N/A & 0.68 & \textbf{0.87} & 0.82 & 0.76 \\
& MMFusion -- fine-tuned & F1 score$\uparrow$ & \textit{0.48} & N/A & 0.66 & \textbf{0.86} & 0.81 & 0.75 \\
\bottomrule
\end{tabular}
}
\end{center}
\end{table*}

\begin{table*}[!t]
\begin{center}
\caption{\new{Spearman's correlation and corresponding p-value between the IFL performance (F1 score), on the one hand, and the generative quality metrics, on the other. All correlation values are relatively low, with p-values equal to 0.000.
}
}
\label{tab:benchmark-quality-corr}
{
\footnotesize
\setlength\tabcolsep{4pt}%
\begin{tabular}{lll cc|cc|ccc}
\toprule
\multirow{2}{*}{IFL Method} &%
\multirow{2}{*}{Version} &%
\multirow{2}{*}{Subsets} &%
\multicolumn{2}{c|}{Aesthetic Quality} & \multicolumn{2}{c|}{ITM} & \multicolumn{3}{c}{Preservation Fidelity} \\
 & & & NIMA & GIQA & ITM & cos & PSNR & SSIM & LPIPS \\
\midrule
TruFor   & original & SP &%
0.36 & 0.42 & -0.31 & 0.07 & N/A & N/A & N/A \\
MMFusion & original & SP &%
0.30 & 0.39  & -0.29 & 0.11 & N/A & N/A & N/A \\
TruFor   & fine-tuned & FR &%
0.28 & 0.34 & -0.03 & 0.23 & 0.24 & 0.28 & -0.32 \\
MMFusion & fine-tuned & FR &%
0.28 & 0.35  & -0.02 & 0.24 & 0.24 & 0.28 & -0.32 \\
\bottomrule
\end{tabular}
}
\end{center}
\end{table*}

\new{
Overall, generative quality metrics do not meaningfully predict forensic localization performance. The detectability differences between SD and FLUX in FR settings therefore cannot be explained by the evaluated quality measures.
However, note that we were limited by the current generative quality metrics.
That is, on the one hand, the aesthetic quality metrics used could only be calculated on the inpainted area, but not on the real-but-regenerated area (as they do not allow masked calculation). On the other hand, it only makes sense to calculate the preservation fidelity metrics on the real-but-regenerated area (and not on the inpainted area).
A more informative direction for future work may be to measure intra-image quality contrast, i.e., discrepancies between inpainted regions and real-but-regenerated regions. Such contrast-based metrics could better capture subtle inconsistencies exploited by forensic localization models, and hence could potentially explain the performance discrepancy between subsets in the fine-tuned FR setting.
}

\subsection{Synthetic Image Detection}\label{sec:benchmark-sid}
We benchmark state-of-the-art SID methods on the TGIF2 FR dataset, which now includes FR subsets generated with FLUX.1 models (in addition to only SD2 and SDXL in the original TGIF dataset), as well as both semantic and random-mask variants.
Note that we do not report results on spliced subsets, since SID methods are not designed to detect local manipulations and hence consistently fail to detect these cases. Instead, in this section, we focus only on the fully regenerated subsets. 
We also expanded the benchmark with four recently introduced methods: SPAI~\cite{karageorgiou2025spai}, SPAI-ITW~\cite{konstantinidou2025navigating}, RINE-ITW~\cite{konstantinidou2025navigating}, and B-Free~\cite{guillaro2024biasfree}. 
Table~\ref{tab:benchmark-sid} reports AUC values for all evaluated methods, where we highlight scores above 0.8 in a bold font. This threshold was chosen empirically to allow for quickly grasping which methods demonstrate decent detection performance.

From the 17 evaluated SID models, 11 demonstrate relatively good performance (AUC $>$ 0.8) on either SD2, SDXL, or both. Specifically, the newly included methods (SPAI, SPAI-ITW, RINE-ITW, and B-Free) all have relatively strong performance on these subsets, with B-Free even scoring 1.00 AUC on both sets.

The evaluation of the new FLUX subsets provides a more challenging picture. 
Out of the 11 SID methods that perform reliably well on SD2 and SDXL, only 4 maintain an AUC above 0.8 across all FLUX subsets. 
In general, all methods experience a performance drop on FLUX, suggesting that its generation process produces fewer or less detectable forensic traces.
For example, B-Free results in AUC values between 0.85 and 0.88 for the FLUX subsets, while it resulted in an AUC value of 1.00 for the SD2 and SDXL subsets.
This finding indicates that FLUX represents a significantly harder detection setting, and highlights the importance of developing SID models that generalize across different families of generative models.

We notice no significant difference in SID performance between semantic and random-mask subsets.

In summary, TGIF2 confirms that modern SID methods can successfully detect fully regenerated images produced by earlier diffusion models such as SD2 and SDXL, but new generative models such as FLUX introduce new challenges and degrade detection performance.
However, it should be stressed that binary classification in SID models are not able to localize generative edits.

\begin{table*}[!t]
\begin{center}
\caption{Evaluation of synthetic image detection methods on the semantic subsets of our TGIF2 Dataset (AUC Score). AUC scores above 0.8 are highlighted in a bold font.
}
\label{tab:benchmark-sid}
{
\footnotesize

\setlength\tabcolsep{1.5pt}%
\begin{tabular}{ll cc|cc|cc|cc|cc|ccc}
\toprule
\multirow{3}{*}{SID Method} & \multirow{3}{*}{Trained on} & 
    \multicolumn{2}{c|}{\multirow{2}{*}{SD2}} & \multicolumn{2}{c|}{\multirow{2}{*}{SDXL}} & \multicolumn{6}{c|}{FLUX.1} & \multicolumn{3}{c}{\multirow{2}{*}{Average}}\\
 & & & & & & \multicolumn{2}{c|}{[schnell]} & \multicolumn{2}{c|}{[dev]} & \multicolumn{2}{c|}{Fill [dev]} \\
 & & Sem & \new{Rand} & Sem & \new{Rand} &
 \new{Sem} & \new{Rand} & \new{Sem} & \new{Rand} & \new{Sem} & \new{Rand} & \new{Sem} & \new{Rand} & \new{All} \\
 \midrule

CNNDetect~\cite{wang2019cnngenerated} & ProGAN & 0.57 & 0.48 & 0.61 & 0.62 & 0.50 & 0.47 & 0.49 & 0.50 & 0.62 & 0.62 & 0.56 & 0.54 & 0.55 \\
Dire~\cite{wang2023dire} & ADM & 0.49 & 0.52 & 0.62 & 0.64 & 0.49 & 0.54 & 0.48 & 0.52 & 0.48 & 0.49 & 0.51 & 0.54 & 0.53 \\
FreqDetect~\cite{frank2020fredetect} & GANs &  0.50 & 0.42 & 0.40 & 0.41 & 0.39 & 0.36 & 0.37 & 0.38 & 0.39 & 0.34 & 0.41 & 0.38 & 0.40 \\
Fusing~\cite{ju2022fusing} & ProGAN & 0.55 & 0.48 & 0.47 & 0.47 & 0.55 & 0.53 & 0.55 & 0.56 & 0.62 & 0.62 & 0.55 & 0.53 & 0.54 \\
NPR~\cite{tan2024NPR} & ProGAN & 0.51 & 0.49 & 0.67 & 0.31 & 0.49 & 0.50 & 0.49 & 0.48 & 0.44 & 0.43 & 0.52 & 0.44 & 0.48 \\
DeFake~\cite{sha2023defake} & Diffusion & 0.62 & 0.60 & 0.61 & 0.59 & 0.67 & 0.66 & 0.70 & 0.67 & 0.63 & 0.61 & 0.65 & 0.63 & 0.64 \\
DIMD~\cite{corvi2023dimd} & LDM & 0.62 & 0.60 & 0.61 & 0.59 & 0.67 & 0.66 & 0.70 & 0.67 & 0.63 & 0.61 & 0.65 & 0.63 & 0.64 \\
UnivFD~\cite{ojha2023univFD} & ProGAN & \textbf{0.82} & \textbf{0.80} & \textbf{0.80} & \textbf{0.81} & 0.63 & 0.60 & 0.58 & 0.59 & 0.69 & 0.71 & 0.70 & 0.70 & 0.70 \\
GramNet~\cite{liu2020gramnet} & GANs & \textbf{0.82} & 0.79 & 0.55 & 0.54 & \textbf{0.81} & \textbf{0.81} & \textbf{0.82} & \textbf{0.82} & \textbf{0.84} & \textbf{0.84} & 0.77 & 0.76 & 0.76 \\
LGrad~\cite{tan2023LGrad} & ProGAN & \textbf{0.85} & \textbf{0.85} & \textbf{0.83} & \textbf{0.80} & \textbf{0.84} & \textbf{0.84} & \textbf{0.85} & \textbf{0.86} & \textbf{0.86} & \textbf{0.84} & \textbf{0.85} & \textbf{0.84} & \textbf{0.84} \\
RINE~\cite{koutlis2024Rine} & ProGAN & \textbf{0.89} & \textbf{0.86} & \textbf{0.89} & \textbf{0.90} & 0.67 & 0.65 & 0.62 & 0.62 & 0.77 & 0.77 & 0.77 & 0.76 & 0.76 \\
 & LDM & \textbf{0.97} & \textbf{0.97} & \textbf{0.93} & \textbf{0.98} & 0.69 & 0.75 & 0.67 & 0.74 & 0.78 & \textbf{0.87} & \textbf{0.81} & \textbf{0.86} & \textbf{0.84} \\
PatchCraft~\cite{zhong2023patchcraft} & ProGAN & \textbf{0.98} & \textbf{0.98} & \textbf{0.95} & \textbf{0.96} & \textbf{0.94} & \textbf{0.94} & \textbf{0.95} & \textbf{0.95} & \textbf{0.92} & \textbf{0.90} & \textbf{0.95} & \textbf{0.95} & \textbf{0.95} \\
SPAI~\cite{karageorgiou2025spai} & LDM & \textbf{0.97} & \textbf{0.96} & 0.77 & 0.78 & \textbf{0.85} & \textbf{0.84} & \textbf{0.86} & \textbf{0.86} & \textbf{0.83} & \textbf{0.81} & \textbf{0.86} & \textbf{0.85} & \textbf{0.85} \\
SPAI-ITW~\cite{konstantinidou2025navigating} & ITW & \textbf{0.81} & 0.75 & \textbf{0.83} & \textbf{0.82} & 0.53 & 0.52 & 0.53 & 0.56 & 0.55 & 0.52 & 0.65 & 0.63 & 0.64 \\
RINE-ITW~\cite{konstantinidou2025navigating} & ITW & \textbf{0.98} & \textbf{0.97} & \textbf{0.98} & \textbf{0.98} & 0.73 & 0.73 & 0.72 & 0.77 & 0.74 & 0.71 & \textbf{0.83} & \textbf{0.83} & \textbf{0.83}\\
B-Free~\cite{guillaro2024biasfree} & LDM & \textbf{1.00} & \textbf{1.00} & \textbf{1.00} & \textbf{0.99} & \textbf{0.86} & 0.78 & \textbf{0.85} & 0.77 & \textbf{0.88} & \textbf{0.82} & \textbf{0.92} & \textbf{0.87} & \textbf{0.90} \\
\bottomrule
\end{tabular}
}
\end{center}
\end{table*}

\subsection{Generative Super-Resolution Attack}\label{sec:benchmark-sr}
This subsection studies the robustness of forensic methods against generative super resolution. Specifically, we apply the open-source Real-ESRGAN~\cite{wang2021realesrgan}, which is one of the most popular super-resolution methods to date (with over $34,000$ stars on GitHub~\cite{diffusers_github_issue}). Additionally, we apply bicubic interpolation as baseline comparison. For the experiments, we apply an increase in resolution with a factor of either 2 or 4, followed by a decrease in resolution with the same factor, in order to obtain an attacked image with the same resolution as the original one.
\new{On average, the PSNR between the bicubic up-and-downsampled image and the original image is 19.2 (regardless of the factor), and the LPIPS is 0.21 and 0.22, for the factor of 2 and 4, respectively.
For the super-resolution method, the PSNR is 18.8 (regardless of the factor), and the LPIPS~\cite{zhang2018lpips} is 0.25 and 0.26, for the factor of 2 and 4, respectively.}
These attacked images are evaluated using both IFL and SID methods, but restricted to the models that demonstrate strong performance in Section~\ref{sec:benchmark-ifl} and~\ref{sec:benchmark-sid}.
Table~\ref{tab:benchmark-superresolution-ifl} and~\ref{tab:benchmark-superresolution-sid} present the results for IFL methods applied on all non-random SP subsets and SID methods applied on all non-random FR subsets, respectively. For each evaluated IFL or SID model, the tables show the average performance on all subsets in addition to the performance and corresponding delta in performance when applying the bicubic and super-resolution attacks. Note that a delta with a negative value means a decrease in performance, and a positive value means an increase in performance.
This allows us to systematically evaluate how super-resolution attacks impact the performance of strong forensic models.

\begin{table*}[!t]
\begin{center}
\caption{Evaluation of super-resolution attack on image forgery localization methods on spliced images of our TGIF2 Dataset, alongside a cubic interpolation attack as baseline (F1 score, along with the $\Delta$ F1 Score).
}
\label{tab:benchmark-superresolution-ifl}
{
\footnotesize
\setlength\tabcolsep{4pt}%
\begin{tabular}{l c cc cc}
\toprule
\multirow{3}{*}{IFL Method} & \multirow{3}{*}{Original} & \multicolumn{4}{c}{Super-resolution method}\\
& & \multicolumn{2}{c}{Cubic interpolation} & \multicolumn{2}{c}{Real-ESRGAN} \\
& & x2 & x4 & x2 & x4 \\
\midrule
CAT-Net~\cite{kwon2022catnetv2} & 0.87 & 0.76 \textsuperscript{-0.11} & 0.73 \textsuperscript{-0.14} & 0.06 \textsuperscript{-0.81} & 0.07 \textsuperscript{-0.80} \\
TruFor~\cite{guillaro2023trufor} & 0.78 & 0.76 \textsuperscript{-0.02} & 0.76 \textsuperscript{-0.02} & 0.21 \textsuperscript{-0.57} & 0.22 \textsuperscript{-0.56}\\
MMFusion~\cite{triaridis2024mmfusion} & 0.69 & 0.65 \textsuperscript{-0.04} & 0.64 \textsuperscript{-0.05} & 0.18 \textsuperscript{-0.51} & 0.51 \textsuperscript{-0.51}\\
\bottomrule
\end{tabular}
}
\end{center}
\end{table*}

\begin{table*}[!t]
\begin{center}
\caption{Evaluation of super-resolution attack on synthetic image detection methods on fully regenerated images in our TGIF2 dataset, alongside a cubic interpolation attack as baseline (AUC Score, along with the $\Delta$ AUC score).
}
\label{tab:benchmark-superresolution-sid}
{
\footnotesize
\setlength\tabcolsep{4pt}%
\begin{tabular}{ll c cc cc}
\toprule
\multirow{3}{*}{SID Method} & \multirow{3}{*}{Trained on} & \multirow{3}{*}{Original} & \multicolumn{4}{c}{Super-resolution method}\\
& & & \multicolumn{2}{c}{Cubic interpolation} & \multicolumn{2}{c}{Real-ESRGAN} \\
& & & x2 & x4 & x2 & x4 \\
\midrule
DIMD~\cite{corvi2023dimd} & LDM & 0.88 & 0.87 \textsuperscript{--0.01} & 0.87 \textsuperscript{--0.01} & 0.75 \textsuperscript{--0.13} & 0.80 \textsuperscript{--0.08} \\
UnivFD~\cite{ojha2023univFD} & ProGAN & 0.70 & 
0.70 \textsuperscript{+0.00} & 0.69 \textsuperscript{--0.01} & 0.68 \textsuperscript{--0.02} & 0.68 \textsuperscript{--0.02}\\
GramNet~\cite{liu2020gramnet} & GANs & 0.76 & 0.76 \textsuperscript{+0.00} & 0.75 \textsuperscript{--0.01} & 0.50 \textsuperscript{--0.26} & 0.50 \textsuperscript{--0.26} \\
LGrad~\cite{tan2023LGrad} & ProGAN & 0.85 & 0.85 \textsuperscript{+0.00} & 0.84 \textsuperscript{--0.01} & 0.49 \textsuperscript{--0.36} & 0.48 \textsuperscript{--0.37} \\
RINE~\cite{koutlis2024Rine} & ProGAN & 0.77 & 0.77 \textsuperscript{+0.00} & 0.76 \textsuperscript{--0.01} & 0.72 \textsuperscript{--0.05} & 0.72 \textsuperscript{--0.05} \\
 & LDM & 0.81 & 0.83 \textsuperscript{+0.02} & 0.83 \textsuperscript{+0.02} & 0.70 \textsuperscript{--0.11} & 0.71 \textsuperscript{--0.10}\\
PatchCraft~\cite{zhong2023patchcraft} & ProGAN & 0.95 & 0.98 \textsuperscript{+0.03} & 0.98 \textsuperscript{+0.03} & 0.53 \textsuperscript{--0.42} & 0.48 \textsuperscript{--0.47} \\
SPAI~\cite{karageorgiou2025spai} & LDM & 0.86 & 0.84 \textsuperscript{--0.02} & 0.84 \textsuperscript{--0.02} & 0.73 \textsuperscript{--0.13} & 0.73 \textsuperscript{--0.13} \\
SPAI--ITW~\cite{konstantinidou2025navigating} & ITW & 0.65 & 0.65 \textsuperscript{+0.00} & 0.66 \textsuperscript{+0.01} & 0.78 \textsuperscript{+0.13} & 0.83 \textsuperscript{+0.18} \\
RINE--ITW~\cite{konstantinidou2025navigating} & ITW & 0.93 & 0.83 \textsuperscript{+0.00} & 0.83 \textsuperscript{+0.00} & 0.81 \textsuperscript{--0.02} & 0.83 \textsuperscript{+0.00} \\
B--Free~\cite{guillaro2024biasfree} & LDM & 0.92 & 0.91 \textsuperscript{--0.01} & 0.91 \textsuperscript{--0.01} & 0.86 \textsuperscript{--0.06} & 0.84 \textsuperscript{--0.08}\\
\bottomrule
\end{tabular}
}
\end{center}
\end{table*}

For IFL methods, Table~\ref{tab:benchmark-superresolution-ifl} shows that super resolution has a pronounced negative impact. 
While bicubic interpolation produces only a modest decrease in F1 scores (i.e., average F1 score decrease between 0.02 and 0.11), applying Real-ESRGAN leads to a substantial drop across all evaluated subsets (i.e., average F1 score decrease between 0.51 and 0.80). 
This indicates that generative SR effectively removes the fine-grained pixel-level traces that IFL methods rely on, making localization of manipulated regions far more difficult.
This is a similar observation as we made for the impact of full regeneration during generative inpainting (in Section~\ref{sec:benchmark-ifl}). Moreover, a similar observation was made in related work, which demonstrated that IFL methods performance drops when applying generative compression using JPEG AI~\cite{cannas2024jpegai}.
This highlights that generative image processing constitutes a significant challenge for current IFL methods, requiring new defenses or adaptations.

\new{Table~\ref{tab:benchmark-superresolution-sid}} shows that the effect of generative super resolution on SID methods is less pronounced than on IFL methods. 
For several models, Real-ESRGAN causes a clear but moderate reduction in AUC, suggesting that their detection cues are also weakened by the generative upscaling process. 
However, other SID models demonstrate much smaller performance drops. In some cases, the performance remains relatively stable (e.g., for UnivFD and RINE-ITW). Notably, for SPAI-ITW, the performance even increases when applying the Real-ESRGAN super-resolution attack.
In comparison, applying the bicubic interpolation (as baseline) does not affect detection performance (approximately 0.0 difference in AUC).
This suggests that some SID approaches rely on more global or semantic inconsistencies that are less affected by generative SR, while others depend on low-level statistical traces that Real-ESRGAN can affect. 

In summary, our experiments show that generative super resolution represents a serious threat to current forensic methods. 
It can erase the traces needed for IFL, and in many cases also moderately harm SID performance, while bicubic interpolation alone does not have such strong effects. 
This highlights the importance of accounting for post-processing operations such as SR in the development of robust forensic methods.

\section{Discussion \& Conclusion}\label{sec:conclusion}
Text-guided inpainting forgeries represent a rapidly evolving and increasingly realistic form of AI-based image editing. 
With TGIF2, we extend our previous dataset (TGIF) by incorporating recent inpainting models, namely the FLUX.1 family, and by introducing random-mask subsets to probe biases in detection methods. Additionally, we extend our previous benchmark by performing fine-tuning to tackle the problem of localization in manipulated but fully regenerated images, as well as evaluating the impact of generative super-resolution methods on forensic methods.

Our experiments demonstrate that existing IFL methods perform well on traditional splicing scenarios, yet some showcase a notable difference in performance between edits with first-generation inpainting methods (i.e., SD and Adobe Firefly) and new-generation methods (i.e., FLUX.1), revealing limitations in their ability to generalize to new inpainting models that significantly increase visual fidelity.
\new{Additionally, we noticed that some IFL methods drop in performance when evaluated on the random, non-semantic subsets, which suggests that these existing methods exhibit a bias towards objects and semantics.}

Moreover, existing IFL models are unable to localize manipulated areas in FR images.
We demonstrated that fine-tuning on FR subsets improves their ability to localize these manipulations, but that it also may introduce biases, which became evident when evaluating on the random-mask subsets. This highlights the importance of training and evaluating methods beyond object-centric manipulations.

\new{We also analyzed the generative quality of the inpainted images, demonstrating that FLUX images are of higher quality. However, we could not establish conclusive relations between generative quality and  IFL performance.}

For SID, our experiments show that several methods perform strongly on SD2 and SDXL, but performance decreases significantly for FLUX subsets. Hence, FLUX generations are harder to detect for existing IFL and SID methods. This emphasizes the need to keep investing in generalizable SID methods, as well as including new generative methods in training sets.

Furthermore, we show that generative super resolution significantly weakens forensic traces and thereby reduces the effectiveness of forensic methods, similar to the fully regenerated manipulations and generative compression~\cite{cannas2024jpegai}. This is especially disruptive in the case of IFL methods, but also moderately affects some SID methods.
\new{In fact, the results suggest that semantics-based SID methods show increased robustness to SR attacks.}

\new{It is interesting to contrast the increased SR-attack robustness of semantics-based SID methods (in Section~\ref{sec:benchmark-sr}) to the revealed semantic-bias vulnerability in (fine-tuned) IFL methods that were exposed in random-mask evaluations (in Section~\ref{sec:benchmark-ifl} and Section~\ref{sec:benchmark-ifl-fine-tune}).
This may seem conflicting, since we demonstrate that semantics is beneficial in certain use cases (e.g., robustness to SR attacks), while an over-reliance on semantic cues can introduce vulnerabilities (e.g., that become apparent under random-mask evaluations).
It should be noted that the use of random, non-semantic masks is not intended to replace object-centric evaluations, but to serve as a stress test that decouples semantic content from manipulation boundaries, enabling a clearer analysis of model bias and cue reliance.
As such, TGIF2 aims to provide complementary evaluation conditions that expose different failure modes. We see semantic reasoning and low-level forensic cues as orthogonal and necessary components, and our benchmark is designed to analyze their trade-offs rather than promote one over the other.}

Future work could focus on expanding the TGIF2 dataset with next-generation AI-based inpainting tools, e.g., those that do not require a user-defined mask, such as FLUX Kontext~\cite{labs2025flux1kontextflowmatching}\new{, FLUX.2~\cite{labs2025flux2}}, and the GPT-4o (ChatGPT) Image Generator~\cite{chen2025gpt4oimage}.
Additionally, a key open challenge is improving the robustness of IFL methods against manipulated images that undergo full regeneration (either during manipulation or after, such as by generative super resolution). Regeneration significantly weakens or destroys forensic traces used by current IFL and SID methods. In this context, it is especially important to closely watch out for potential biases.

In summary, TGIF2 provides an updated benchmark that captures the latest advances in text-guided inpainting, highlights strengths and weaknesses of state-of-the-art forensic methods and their fine-tuned versions, and contributes new challenging subsets that will help the community to track progress against this continuously evolving threat.

\backmatter

\bmhead{Acknowledgements}
Not applicable.

\section*{Declarations}

\begin{description}
\item[\textbf{Funding}]
This work was funded by the Flemish government’s Department of Culture, Youth \& Media (under the COM-PRESS project), by IDLab (Ghent University -- imec), by Flanders Innovation \& Entrepreneurship (VLAIO), by Research Foundation -- Flanders (FWO) (V419524N \& G0A2523N), and by the European Union under the Horizon Europe projects AI4Trust (grant number 101070190) and AI-CODE (grant number 101135437).
\item [\textbf{Competing interests}]
The authors declare that they have no competing interests.
\item [\textbf{Ethics approval and consent to participate}]
Not applicable.
\item [\textbf{Consent for publication}]
The authors give consent for publication.
\item [\textbf{Data availability}]
\new{The TGIF2 dataset is available at \url{https://github.com/IDLabMedia/tgif-dataset}.}
\item [\textbf{Materials availability}]
\new{The TGIF2 materials are available at \url{https://github.com/IDLabMedia/tgif-dataset}.}
\item [\textbf{Code availability}]
\new{The TGIF2 code are available at \url{https://github.com/IDLabMedia/tgif-dataset}.}
\item [\textbf{Author contribution}]
HM: dataset preparation, initial experiments, interpretation of experiments, main manuscript text writing.
DK: large-scale benchmark experiments, interpretation of experiments.
PG: large-scale benchmark experiments, interpretation of experiments.
SP: guidance, funding acquisition.
PL: guidance, funding acquisition.
GVW: guidance, funding acquisition.
All authors revised and approved the manuscript.
\end{description}

\begin{appendices}
\section{\new{Image Forgery Localization -- IoU}}
\label{app:ifl-iou}

\new{The IoU results for IFL on the spliced subsets are given in Table~\ref{tab:benchmark-ifl-sp-iou} (i.e., corresponding to the F1 results in Table~\ref{tab:benchmark-ifl-sp}). The IoU results for the fully regenerated subsets are given in Table~\ref{tab:benchmark-ifl-fr-iou} (i.e., corresponding to the F1 results in Table~\ref{tab:benchmark-ifl-fr}) for the original IFL methods, and in Table~\ref{tab:benchmark-ifl-fine-tune-iou} (i.e., corresponding to the F1 results in Table~\ref{tab:benchmark-ifl-fine-tune}), for the fine-tuned IFL methods.}

\begin{table*}[!htb]
\begin{center}
\caption{\new{Evaluation of image forgery localization methods on the \textbf{spliced (SP)} subsets of our TGIF2 dataset, for both the semantic (sem) and random (rand) versions (\textbf{IoU}). IoU values above 0.6 are highlighted in a bold font.
Just as in Table~\ref{tab:benchmark-ifl-sp}, some IFL methods (i.e., CAT-Net, TruFor \& MMFusion) perform well on SP subsets.}
}
\label{tab:benchmark-ifl-sp-iou}
{
\footnotesize
\setlength\tabcolsep{3pt}%
\begin{tabular}{l cc|c|cc|cc|cc|ccc}
\toprule
\multirow{3}{*}{IFL Method}
 & \multicolumn{2}{c|}{\multirow{2}{*}{SD2}} & \multirow{2}{*}{PS} & \multicolumn{6}{c|}{FLUX.1} & \multicolumn{3}{c}{\multirow{2}{*}{Average}} \\
 & & & & \multicolumn{2}{c|}{[schnell]} & \multicolumn{2}{c|}{[dev]} & \multicolumn{2}{c|}{Fill [dev]} \\
 & Sem & Rand & Sem &
 Sem & Rand & Sem & Rand & Sem & Rand & Sem & Rand & All \\
\midrule
PSCC-Net~\cite{liu2022pscc} & 0.12 & 0.13 & 0.31 & 0.02 & 0.06 & 0.01 & 0.03 & 0.06 & 0.17 & 0.10 & 0.10 & 0.10 \\
SPAN~\cite{hu2020span}  & 0.00 & 0.00 & 0.00 & 0.00 & 0.00 & 0.00 & 0.00 & 0.00 & 0.01 & 0.00 & 0.00  & 0.00 \\
ImageForensicsOSN~\cite{wu2022ImageForensicsOSN} & 0.16 & 0.05 & 0.25 & 0.22 & 0.07 & 0.16 & 0.04 & 0.08 & 0.07 & 0.18 & 0.06 & 0.12 \\
MVSS-Net++~\cite{dong2022mvss} & 0.05 & 0.02 & 0.05 & 0.11 & 0.04 & 0.08 & 0.02 & 0.06 & 0.07 & 0.07 & 0.04 & 0.06 \\
Mantranet~\cite{wu2019mantranet} & 0.10 & 0.09 & 0.44 & 0.04 & 0.01 & 0.03 & 0.01 & 0.03 & 0.02 & 0.13 & 0.03 & 0.08 \\
CAT-Net~\cite{kwon2022catnetv2} & \textbf{0.81} & \textbf{0.87} & \textbf{0.77} & \textbf{0.81} & \textbf{0.88} & \textbf{0.78} & \textbf{0.87} & \textbf{0.79} & \textbf{0.89} & \textbf{0.79} & \textbf{0.88} & \textbf{0.83} \\
TruFor~\cite{guillaro2023trufor} & \textbf{0.74} & \textbf{0.79} & \textbf{0.69} & \textbf{0.69} & \textbf{0.60} & \textbf{0.61} & 0.48 & \textbf{0.64} & \textbf{0.63} & 0.67 & 0.63 & 0.65 \\
MMFusion~\cite{triaridis2024mmfusion} & \textbf{0.62} & \textbf{0.63} & \textbf{0.62} & \textbf{0.60} & 0.50 & 0.49 & 0.37 & 0.54 & 0.50 & 0.57 & 0.50 & 0.54 \\
\bottomrule
\end{tabular}
}
\end{center}
\end{table*}

\begin{table*}[!htb]
\begin{center}
\caption{\new{Evaluation of image forgery localization methods on the \textbf{fully regenerated (FR)} subsets of our TGIF2 dataset, for both the semantic (sem) and random (rand) versions (\textbf{IoU}).
The last rows show the IFL methods fine-tuned on All FR Sem subsets.
Just as in Table~\ref{tab:benchmark-ifl-fr}, the performance is low for all original IFL methods on FR subsets, while it is significantly higher for the fine-tuned variants.}
}
\label{tab:benchmark-ifl-fr-iou}
{
\footnotesize
\setlength\tabcolsep{2pt}%
\begin{tabular}{l cc|cc|cc|cc|cc|ccc}
\toprule
\multirow{3}{*}{IFL Method}
  & \multicolumn{2}{c|}{\multirow{2}{*}{SD2}} & \multicolumn{2}{c|}{\multirow{2}{*}{SDXL}} & \multicolumn{6}{c|}{FLUX.1} & \multicolumn{3}{c}{\multirow{2}{*}{Average}} \\
 & & & & & \multicolumn{2}{c|}{[schnell]} & \multicolumn{2}{c|}{[dev]} & \multicolumn{2}{c|}{Fill [dev]} \\
 & Sem & Rand & Sem & Rand &
 Sem & Rand & Sem & Rand & Sem & Rand & Sem & Rand & All \\
\midrule
PSCC-Net~\cite{liu2022pscc} %
& 0.03 & 0.02 & 0.03 & 0.05 & 0.01 & 0.01 & 0.01 & 0.01 & 0.02 & 0.03 & 0.02 & 0.03 & 0.02 \\
SPAN~\cite{hu2020span} %
 & 0.00 & 0.00 & 0.00 & 0.00 & 0.00 & 0.00 & 0.00 & 0.00 & 0.00 & 0.00 & 0.00 & 0.00 & 0.00 \\
ImageForensicsOSN~\cite{wu2022ImageForensicsOSN} %
 & 0.14 & 0.05 & 0.12 & 0.05 & 0.22 & 0.05 & 0.16 & 0.04 & 0.06 & 0.03 & 0.14 & 0.04 & 0.09 \\
MVSS-Net++~\cite{dong2022mvss} %
 & 0.04 & 0.02 & 0.06 & 0.03 & 0.11 & 0.02 & 0.09 & 0.02 & 0.05 & 0.03 & 0.07 & 0.02 & 0.05 \\
Mantranet~\cite{wu2019mantranet} %
 & 0.02 & 0.00 & 0.03 & 0.03 & 0.03 & 0.01 & 0.02 & 0.01 & 0.01 & 0.01 & 0.02 & 0.01 & 0.02 \\
CAT-Net~\cite{kwon2022catnetv2} %
 & 0.03 & 0.01 & 0.02 & 0.00 & 0.04 & 0.01 & 0.03 & 0.00 & 0.02 & 0.01 & 0.03 & 0.01 & 0.02 \\
TruFor~\cite{guillaro2023trufor} %
 & 0.13 & 0.04 & 0.12 & 0.06 & 0.25 & 0.05 & 0.18 & 0.04 & 0.10 & 0.05 & 0.16 & 0.05 & 0.10 \\
MMFusion~\cite{triaridis2024mmfusion} %
 & 0.11 & 0.03 & 0.13 & 0.06 & 0.26 & 0.05 & 0.15 & 0.02 & 0.08 & 0.04 & 0.15 & 0.04 & 0.09 \\
\midrule
TruFor -- fine-tuned %
 & 0.37 & 0.28 & 0.58 & \textbf{0.75} & \textbf{0.80} & \textbf{0.90} & \textbf{0.74} & \textbf{0.82} & \textbf{0.67} & \textbf{0.78} & \textbf{0.63} & \textbf{0.71} & \textbf{0.67} \\
MMFusion -- fine-tuned %
 & 0.37 & 0.29 & 0.55 & \textbf{0.75} & \textbf{0.78} & \textbf{0.89} & \textbf{0.73} & \textbf{0.80} & \textbf{0.66} & \textbf{0.77} & \textbf{0.62} & \textbf{0.70} & \textbf{0.66} \\
\bottomrule
\end{tabular}
}
\end{center}
\end{table*}

\begin{table*}[!htb]
\begin{center}
\caption{\new{Evaluation of \textbf{individually fine-tuned} image forgery localization methods on the \textbf{fully regenerated (FR)} subsets of our TGIF2 dataset, for both the semantic (sem) and random (rand) versions (\textbf{IoU}).
IoU values above 0.6 are highlighted in a bold font.
Just as in Table~\ref{tab:benchmark-ifl-fine-tune}, we notice good in-domain performance but bad out-of-domain generalization, as well as a potential bias towards semantics.
}
}
\label{tab:benchmark-ifl-fine-tune-iou}
{
\footnotesize
\setlength\tabcolsep{1pt}%
\begin{tabular}{lll cc|cc|cc|cc|cc|ccc}
\toprule
\multirow{3}{*}{IFL Method} & \multicolumn{2}{c}{\multirow{3}{*}{Fine-tune set}} &
 \multicolumn{2}{c|}{\multirow{2}{*}{SD2}} & \multicolumn{2}{c}{\multirow{2}{*}{SDXL}} & \multicolumn{6}{c|}{FLUX.1} &  \multicolumn{3}{c}{\multirow{2}{*}{Average}} \\
 & & & & & & & \multicolumn{2}{c|}{[schnell]} & \multicolumn{2}{c|}{[dev]} & \multicolumn{2}{c|}{Fill [dev]} \\
 & & & Sem & Rand & Sem & Rand &
 Sem & Rand & Sem & Rand & Sem & Rand & Sem & Rand & All \\
 \midrule

\multirow{9}{*}{TruFor}%
   & \multirow{3}{*}{All} & Sem+Rand %
 & 0.37 & 0.28 & 0.58 & \textbf{0.75} & \textbf{0.80} & \textbf{0.90} & \textbf{0.74} & \textbf{0.82} & \textbf{0.67} & \textbf{0.78} & \textbf{0.63} & \textbf{0.71} & \textbf{0.67} \\
   & & Sem %
 & 0.21 & 0.29 & 0.41 & \textbf{0.71} & \textbf{0.60} & \textbf{0.87} & 0.53 & \textbf{0.78} & 0.53 & \textbf{0.74} & 0.55 & 0.41 & 0.48 \\
   & & Rand %
 & 0.38 & 0.15 & 0.50 & 0.52 & \textbf{0.71} & 0.52 & 0.59 & 0.37 & 0.55 & 0.50 & 0.46 & \textbf{0.68} & 0.57 \\[6pt]
 
  & \multirow{2}{*}{SD2} & Sem %
 & 0.38 & 0.14 & 0.20 & 0.11 & 0.29 & 0.11 & 0.22 & 0.05 & 0.15 & 0.06 & 0.25 & 0.09 & 0.17 \\
   &  & Rand %
 & 0.21 & 0.29 & 0.11 & 0.14 & 0.08 & 0.06 & 0.05 & 0.03 & 0.08 & 0.08 & 0.11 & 0.12 & 0.11 \\
    & \multirow{2}{*}{SDXL} & Sem %
 & 0.07 & 0.01 & 0.55 & \textbf{0.61} & 0.14 & 0.01 & 0.06 & 0.01 & 0.04 & 0.01 & 0.17 & 0.13 & 0.15 \\
    &  & Rand %
 & 0.01 & 0.00 & 0.41 & \textbf{0.73} & 0.01 & 0.00 & 0.00 & 0.01 & 0.02 & 0.02 & 0.09 & 0.15 & 0.12 \\
     & \multirow{2}{*}{Flux} & Sem %
 & 0.07 & 0.01 & 0.14 & 0.03 & \textbf{0.83} & \textbf{0.85} & \textbf{0.79} & \textbf{0.74} & \textbf{0.70} & \textbf{0.72} & 0.51 & 0.47 & 0.49 \\
    &  & Rand %
 & 0.01 & 0.01 & 0.03 & 0.04 & \textbf{0.65} & \textbf{0.91} & \textbf{0.61} & \textbf{0.86} & 0.58 & \textbf{0.77} & 0.38 & 0.52 & 0.45 \\
 
 \midrule
 
  \multirow{9}{*}{MMFusion}%
  & \multirow{3}{*}{All} & Sem+Rand %
 & 0.37 & 0.29 & 0.55 & \textbf{0.75} & \textbf{0.78} & \textbf{0.89} & \textbf{0.73} & \textbf{0.80} & \textbf{0.66} & \textbf{0.77} & \textbf{0.62} & \textbf{0.70} & 0.66 \\
 &  & Sem %
 & 0.36 & 0.11 & 0.47 & 0.50 & \textbf{0.70} & 0.48 & \textbf{0.62} & 0.41 & 0.59 & 0.57 & 0.55 & 0.42 & 0.48 \\
 &  & Rand %
 & 0.18 & 0.30 & 0.39 & \textbf{0.71} & 0.54 & \textbf{0.85} & 0.49 & \textbf{0.75} & 0.52 & \textbf{0.72} & 0.42 & \textbf{0.67} & 0.55 \\[6pt]
 
  & \multirow{2}{*}{SD2} & Sem %
 & 0.40 & 0.15 & 0.19 & 0.12 & 0.26 & 0.10 & 0.19 & 0.05 & 0.12 & 0.06 & 0.23 & 0.10 & 0.16 \\
   &  & Rand %
 & 0.20 & 0.24 & 0.11 & 0.12 & 0.05 & 0.03 & 0.04 & 0.02 & 0.06 & 0.05 & 0.09 & 0.09 & 0.09 \\
    & \multirow{2}{*}{SDXL} & Sem %
 & 0.06 & 0.01 & 0.55 & \textbf{0.60} & 0.10 & 0.01 & 0.04 & 0.01 & 0.03 & 0.01 & 0.16 & 0.13 & 0.14 \\
    &  & Rand %
 & 0.01 & 0.01 & 0.41 & \textbf{0.71} & 0.02 & 0.00 & 0.01 & 0.01 & 0.03 & 0.03 & 0.10 & 0.15 & 0.12 \\
     & \multirow{2}{*}{Flux} & Sem %
 & 0.07 & 0.01 & 0.14 & 0.03 & \textbf{0.83} & \textbf{0.80} & \textbf{0.77} & \textbf{0.68} & \textbf{0.69} & \textbf{0.71} & 0.50 & 0.45 & 0.47 \\
    &  & Rand %
 & 0.02 & 0.01 & 0.04 & 0.05 & \textbf{0.65} & \textbf{0.90} & \textbf{0.61} & \textbf{0.85} & \textbf{0.60} & \textbf{0.77} & 0.38 & 0.52 & 0.45 \\
 
\bottomrule
\end{tabular}
}
\end{center}
\end{table*}

\end{appendices}

\newpage

\bibliography{refs}

\end{document}